\documentclass[sigconf]{acmart}
\AtBeginDocument{%
  }
\setcopyright{acmlicensed}
\copyrightyear{2025}
\acmYear{2025}
\setcopyright{acmlicensed}\acmConference[MM '25]{Proceedings of the 33rd
ACM International Conference on Multimedia}{October 27--31, 2025}{Dublin,
Ireland}
\acmBooktitle{Proceedings of the 33rd ACM International Conference on
Multimedia (MM '25), October 27--31, 2025, Dublin, Ireland}
\acmDOI{10.1145/3746027.3755612}
\acmISBN{979-8-4007-2035-2/2025/10}

\usepackage{epsfig}
\usepackage{graphicx}
\usepackage{multirow}
\usepackage{float}
\usepackage{pifont}
\usepackage{booktabs}
\usepackage{caption}
\usepackage{color}
\usepackage{colortbl}
\usepackage[font=small]{caption}

\usepackage{amsmath,amsfonts,bm}

\def\eqref#1{equation~\ref{#1}}

\def\1{\bm{1}}

\DeclareMathAlphabet{\mathsfit}{\encodingdefault}{\sfdefault}{m}{sl}
\SetMathAlphabet{\mathsfit}{bold}{\encodingdefault}{\sfdefault}{bx}{n}

\newcommand{\bB}{\mathbf{B}}
\newcommand{\bc}{\mathbf{c}}\newcommand{\bC}{\mathbf{C}}

\newcommand{\bF}{\mathbf{F}} 
\newcommand{\bG}{\mathbf{G}}
\newcommand{\bH}{\mathbf{H}}

\newcommand{\bo}{\mathbf{o}}
\newcommand{\bP}{\mathbf{P}}
\newcommand{\bq}{\mathbf{q}}
\newcommand{\bR}{\mathbf{R}}
\newcommand{\bS}{\mathbf{S}}
\newcommand{\bT}{\mathbf{T}}

\newcommand{\bx}{\mathbf{x}}

\newcommand{\bmu}{\boldsymbol{\mu}}

\newcommand{\bOmega}{\boldsymbol{\Omega}}

\newcommand{\nR}{\mathbb{R}}

\newcommand{\cL}{\mathcal{L}}

\newcommand{\figrefer}[1]{Figure~\ref{#1}}
\newcommand{\eqnref}[1]{Eq.~\ref{#1}}
\newcommand{\tabnref}[1]{Table~\ref{#1}}

\usepackage{xspace}
\makeatletter
\DeclareRobustCommand\onedot{\futurelet\@let@token\@onedot}
\def\@onedot{\ifx\@let@token.\else.\null\fi\xspace}

\def\etc{{etc}\onedot}

\makeatother

\usepackage{balance}
\usepackage{hyperref}
\usepackage{url}
\usepackage{wrapfig}
\usepackage{pifont}
\newcommand{\cmark}{\ding{51}} %
\newcommand{\xmark}{\ding{55}} %
\begin{document}

\title{Casual3DHDR: High Dynamic Range 3D Gaussian Splatting from casually-captured Videos}

\author{Shucheng Gong}
\authornote{These authors contributed equally to this research.}
\authornote{Work was done as an intern in Westlake University.}
\email{shuchenggong@whu.edu.cn}
\orcid{0009-0001-6964-8257}
\affiliation{%
  \institution{Westlake University}
  \city{Hangzhou}
  \country{China}
}
\affiliation{%
  \institution{Wuhan University}
  \city{Wuhan}
  \country{China}
}

\author{Lingzhe Zhao}
\authornotemark[1]
\email{zhaolingzhe@westlake.edu.cn}
\orcid{0009-0005-8000-1525}
\affiliation{%
  \institution{Westlake University}
  \city{Hangzhou}
  \country{China}
}

\author{Wenpu Li}
\email{liwenpu@westlake.edu.cn}
\orcid{0009-0009-6794-1810}
\authornotemark[1]
\affiliation{%
  \institution{Westlake University}
  \city{Hangzhou}
  \country{China}}

\author{Hong Xie}
\authornote{Co-corresponding authors}
\email{hxie@sgg.whu.edu.cn}
\orcid{0000-0002-0956-0421}

\affiliation{%
  \institution{Wuhan University}
  \city{Wuhan}
  \country{China}
}

\author{Yin Zhang}
\email{zhangyin@westlake.edu.cn}
\orcid{0000-0002-9347-0241}
  
\affiliation{%
\institution{Zhejiang University}
  \institution{Westlake University}
  \city{Hangzhou}
  \country{China}
}

\author{Shiyu Zhao}
\orcid{0000-0003-3098-8059}
\email{zhaoshiyu@westlake.edu.cn}
\affiliation{%
  \institution{Westlake University}
  \city{Hangzhou}
  \country{China}
}

\author{Peidong Liu}
\orcid{0000-0002-9767-6220}
\email{liupeidong@westlake.edu.cn}
\authornotemark[3]
\affiliation{%
  \institution{Westlake University}
  \city{Hangzhou}
  \country{China}
}

\renewcommand{\shortauthors}{Shucheng Gong et al.}

\begin{abstract}
Photo-realistic novel view synthesis from multi-view images, such as neural radiance field (NeRF) and 3D Gaussian Splatting (3DGS), has gained significant attention for its superior performance. However, most existing methods rely on low dynamic range (LDR) images, limiting their ability to capture detailed scenes in high-contrast environments. While some prior works address high dynamic range (HDR) scene reconstruction, they typically require multi-view sharp images with varying exposure times captured at fixed camera positions, which is time-consuming and impractical. To make data acquisition more flexible, we propose \textbf{Casual3DHDR}, a robust one-stage method that reconstructs 3D HDR scenes from casually-captured auto-exposure (AE) videos, even under severe motion blur and unknown, varying exposure times. Our approach integrates a continuous-time camera trajectory into a unified physical imaging model, jointly optimizing exposure times, camera trajectory, and the camera response function (CRF). Extensive experiments on synthetic and real datasets show that \textbf{Casual3DHDR} outperforms existing methods in robustness and rendering quality.
\end{abstract}



\begin{teaserfigure}
  \includegraphics[width=\textwidth]{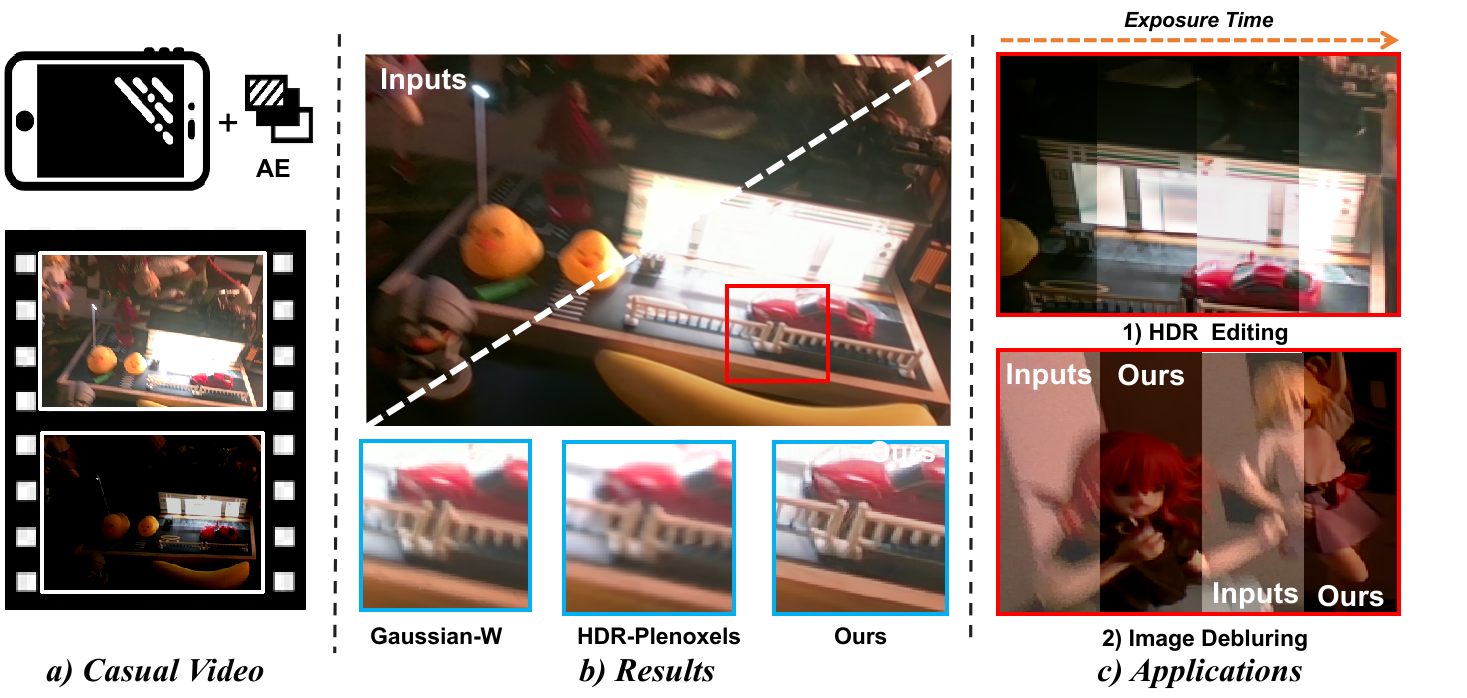}
  \caption{ a) Our method can reconstruct 3D HDR scenes from videos casually-captured with AE enabled. b) Our approach achieves superior rendering quality compared to methods like Gaussian-W and HDR-Plenoxels. c) After 3D HDR reconstruction, we can not only synthesize novel view, but also perform various downstream tasks, such as 1) HDR exposure editing, 2) Image deblurring.}
  \Description{Enjoying the baseball game from the third-base
  seats. Ichiro Suzuki preparing to bat.}
  \label{fig:teaser}
\end{teaserfigure}

\maketitle


\section{Introduction}

\label{sec:intro}

Photo-realistic Novel View Synthesis (NVS) is a vital area in computer vision, enabling applications such as virtual reality/augmented reality (VR/AR), autonomous driving, and embodied AI by providing immersive visual experiences and advanced perception capabilities. Neural Radiance Fields (NeRFs)~\cite{mildenhall2020nerf} and 3D Gaussian Splatting (3DGS)~\cite{kerbl3Dgaussians}, have gained prominence in NVS due to their high-quality 3D scene rendering, inspiring a wealth of subsequent research~\cite{mueller2022instant,zhao2024badgaussians,wang2023badnerf,huang2022hdr}.

However, most NVS methods rely on well-exposed, low dynamic range (LDR) images with consistent lighting, struggling to handle high dynamic range (HDR) scenes where details in very bright or dark regions are lost. This limitation restricts the reconstruction of fine details in HDR environments, reducing NVS applicability in real-world scenarios. While 2D HDR content has been standardized and widely adopted~\cite{hannuksela2015high,bt2100,alakuijala2019jpeg}, 3D HDR free-viewpoint content remains an emerging field with significant potential. Thus, reconstructing 3D HDR scenes is essential for improving visual fidelity and supporting downstream tasks.

Current 3D HDR NVS methods can be divided into two categories.
The first category, e.g. RawNeRF~\citep{mildenhall2022rawnerf} and LE3D~\citep{jin2024le3d} \etc, takes in noisy RAW images, aiming to reconstruct noise-free 3D HDR scenes. The second category, represented by \citep{huang2022hdr, cai2024hdrgsefficienthighdynamic, jun2022hdr, wang2024cinematic,wu2024hdrgshighdynamicrange}, draws inspiration from HDR imaging (HDRI), using multi-exposure LDR images as inputs to reconstruct the 3D HDR scene while learning camera response function (CRF). However, the strict inputs limit their flexibility and broader applications. The challenges include:
\begin{itemize}
    \item Data acquisition of RAW images and accurate exposure time is usually expensive due to the use of professional equipment;
    \item In low-light conditions, long exposure times increase the risk of motion blur from camera shake, reducing reconstruction quality;
    \item The geometric consistency will be compromised if given inaccurate camera pose initialization, as the camera poses are not being optimized.
\end{itemize}

Thus, a key challenge is reducing the cost of data acquisition, enabling high-quality and robust 3D HDR scene reconstruction with consumer-grade devices.

Most modern consumer-grade cameras use auto-exposure (AE) during video recording, automatically adjusting exposure time based on ambient lighting. This expands the captured dynamic range in the video, making it possible to reconstruct 3D HDR scenes. However, naively applying these videos to existing HDR 3D reconstruction methods presents several challenges:

\begin{itemize}
\item Although some cameras support providing exposure time in images stored in EXIF metadata, it is often impossible to obtain the exposure time for each frame in a video format;
\item AE can cause inconsistencies in brightness between frames, leading to pose estimation errors in structure from motion (SfM) frameworks;
\item Videos suffer from severe motion blur due to the camera movement during longer exposure times.
\end{itemize}
Addressing these issues requires modeling camera motion during exposure, rather than assuming a static camera as in prior work. Motion blur and brightness variations both depend on exposure time: longer exposures increase blur and brightness. Previous studies~\cite{huang2022hdr,wang2023badnerf} have decoupled these effects into camera movement and irradiance accumulation. Building on this, we introduce a continuous-time trajectory on $\mathbb{SE}(3)$ to represent video camera motions, enabling joint optimization of poses and exposure times. We also design a CRF module to convert accumulated irradiance (HDR) into LDR image brightness. By integrating camera motion into a unified differentiable rendering framework, we jointly optimize the trajectory, CRF, and exposure times.

In summary, our {\bf{contributions}} are as follows:

\begin{itemize}
	\item \textbf{Casual3DHDR}: A one-stage pipeline build upon a unified imaging model that leverages a continuous-time camera trajectory to jointly optimize poses, CRF, exposure times, and 3D HDR representation, enabling low-cost reconstruction from casually-captured auto-exposure (AE) videos.
	\item A dataset of synthetic and real-world videos with significant brightness variations and motion blur, supporting further research.
	\item Extensive experiments demonstrating SOTA performance in reconstructing high-quality HDR scenes from casual videos.
\end{itemize}

\section{Related Works}

\subsection{High Dynamic Range Imaging}

High Dynamic Range Imaging (HDRI) enhances luminosity beyond standard digital imaging by merging multi-exposure LDR images. In video capture, alternating long and short exposures achieves similar effects. Recently, deep learning approaches have treated HDRI as an image domian translation task, designing networks to convert LDR to HDR images. However, camera motion during exposure time often lead to ghosting artifacts. To address this, \cite{CGF:Gryad:15} proposed an adaptive metering algorithm to adjust exposure and reduce motion artifacts, while other methods use spatial attention to mitigate motion blur. With the advent of 3D scene representations such as NeRF and 3DGS, methods like \citep{huang2022hdr,jun2022hdr,cai2024hdrgsefficienthighdynamic,huang2024ltm,wang2024cinematic} have emerged to reconstruct 3D HDR scenes and calibrate CRFs simultaneously. While these methods are effective, they often rely on precise exposure times and struggle with motion blur, highlighting the need for improved robustness.

\subsection{Image deblurring}
Image deblurring aims to restore sharp images from blurred ones and can be categorized into three types. The first type uses hand-crafted priors, like total variation and heavy-tailed gradient priors, to solve for the blur kernel \citep{NIPS2009_3dd48ab3,cho2009fast}, but struggles with different kernels producing similar effects. The second type, deep learning-based, achieves end-to-end restoration using large datasets \citep{Zamir2021MPRNet, Tsai2022Stripformer}. The third type leverages multi-view blurred images to reconstruct the 3D scene and deblur the images with geometric constraints. Works like \citep{ma2021deblur, Peng2022PDRF, lee2023dp} jointly learn the blur kernel and radiance field, while BAD-NeRF~\citep{wang2023badnerf} introduces a physical motion blur model that jointly optimizes the radiance field and camera trajectories. Following BAD-NeRF~\citep{wang2023badnerf}, emerging works \citep{lee2023exblurf, li2024usb, lee2024smurf, lee2024crim, sun2024_dyblurf, chen2024deblur, oh2024deblurgs, zhao2024badgaussians, yu2024evagaussians, li2024BeNeRF, qi2024ebadnerf, tang2024lse} have demonstrated the effectiveness of using $\mathbb{SE}(3)$ trajectory representations for joint 3D reconstruction and blur image recovery.

\subsection{Robust Novel View Synthesis}

Novel view synthesis aims to generate images from arbitrary viewpoints using input images with known poses. Neural Radiance Fields (NeRF)\citep{martinbrualla2020nerfw} and 3D Gaussian Splatting (3DGS)\citep{kerbl3Dgaussians} advanced this field by reconstructing  3D scene as radiance field and achieve great success.

Most synthesis methods assume high-quality input data, but quality degrades significantly with blurry images, large exposure variations, or inaccurate poses. NeRF-W~\citep{martinbrualla2020nerfw} and Gaussian-W~\citep{zhang2024gaussian} address these challenges by adding optimizable appearance vectors to model varied conditions. HDR-NeRF~\citep{huang2022hdr} and HDR-Plenoxels~\citep{jun2022hdr} reconstruct HDR scenes even with dynamic exposure variations. Other methods, like \cite{Fu_2024_CVPR}, handle pose inaccuracies, while BAD-Gaussians~\citep{zhao2024badgaussians} incorporates camera motion models to deblur inputs during reconstruction.$I^2$-SLAM~\citep{bae2024i2slaminvertingimagingprocess} is a concurrent approach capable of processing images with exposure inconsistencies and blur, though it focuses on RGB-D SLAM and has different representations of trajectory and CRF. These methods collectively improve robustness for NVS under challenging conditions.

\section{Methods}

\label{sec:method}

In this section, we will provide a detailed explanation of our proposed {\bf Casual3DHDR}, which takes an AE video as input. 
We first use continuous-time trajectory to represent the camera motion during video recording rather than addressing each camera pose individually, so that the camera pose and exposure time can be jointly optimized through bundle adjustment. Subsequently, we design a camera response function (CRF) module which can convert the accumulation of irradiance (HDR image) over the exposure time to pixel value (LDR image). Due to the fact that the exposure time is the connection between the continuous-time trajectory and CRF module, we can treat them as a unified differentiable imaging  model thus the trajectory, CRF and exposure time can be jointly optimized and constrained together.

\subsection{Preliminary: 3D Gaussian Splatting}
\label{preliminary}
3DGS represents the scene as 3D Gaussian primitives denoted as $\bG$. Each primitive is characterized by a mean position $\boldsymbol{\mu} \in \mathbb{R}^3$, opacity $\bo \in \mathbb{R}$, color $\bc \in \mathbb{R}^3$, and a 3D covariance matrix $\mathbf{\Sigma} \in \mathbb{R}^{3 \times 3}$. To ensure $\mathbf{\Sigma}$ remains positive semi-definite, it is parameterized by a scaling matrix $\bS \in \mathbb{R}^3$ and a rotation matrix $\bR \in \mathbb{R}^{3 \times 3}$ stored as a quaternion $\bq \in \mathbb{R}^4$. During rendering, the 3D Gaussians are projected onto the image plane at a given pose $\bP_i = \{\mathbf{R}_c \in \mathbb{R}^{3 \times 3}, \mathbf{t}_c \in \mathbb{R}^3\}$ , transforming $\mathbf{\Sigma}$ into a 2D covariance matrix $\mathbf{\Sigma^{\prime}} \in \mathbb{R}^{2 \times 2}$. These can be mathematically expressed as:
\begin{equation}
    \bG(\bx) = e^{-\frac{1}{2}(\bx-\bmu)^{\top}\mathbf{\Sigma}^{-1}(\bx-\bmu)}
    \label{eq:gauss}
\end{equation}
\begin{equation}
    \mathbf{\Sigma} = \mathbf{R} \mathbf{S} \mathbf{S}^{T} \mathbf{R}^{T},\mathbf{\Sigma^{\prime}} = \mathbf{J} \mathbf{R}_c \mathbf{\Sigma} \mathbf{R}_c^T \mathbf{J}^{T},
    \label{eq:covariance3d}
\end{equation}

where $\mathbf{J} \in \mathbb{R}^{2 \times 3}$ is the Jacobian of the affine approximation of the projective transformation. Next, the 2D Gaussians undergo depth sorting followed by tile-based rasterization. The final color values for individual pixels are obtained using $\alpha$-blending:
\vspace{-0.7em}
\begin{equation}
    \bC(x,y,\bP_i) = \sum_{i=1}^{N} \bc_i \alpha_i \prod_{j=1}^{i-1} (1-\alpha_j)
    \label{eq:render}
\end{equation}
\vspace{-1em}
\begin{equation}
    \alpha_i = \bo_i \cdot \exp(-\sigma_i), \quad
    \sigma_i = \frac{1}{2} \Delta_i^T \mathbf{\Sigma^{\prime}}^{-1} \Delta_i,
    \label{eq:a-blender}
\end{equation}
where $\bc_i$ is the learnable color of each Gaussian, and $\alpha_i$ is the alpha value determined by the 2D covariance $\mathbf{\Sigma^{\prime}}$ multiplied by the learned opacity $\bo$. ${\rm \Delta}_i \in \mathbb{R}^2$ represents the offset between the pixel center and the 2D Gaussian center.
The above derivations show that the rendered pixel color, $\bC$ in Eq.(\ref{eq:render}), is differentiable with respect to all learnable parameters $\bG$ and camera poses $\bP_i$, which is crucial for our bundle adjustment formulation.

\begin{figure*}[!t]

	\small
	\begin{center}
		\setlength{\belowcaptionskip}{-1.6em}
		\includegraphics[width=1.0\textwidth]{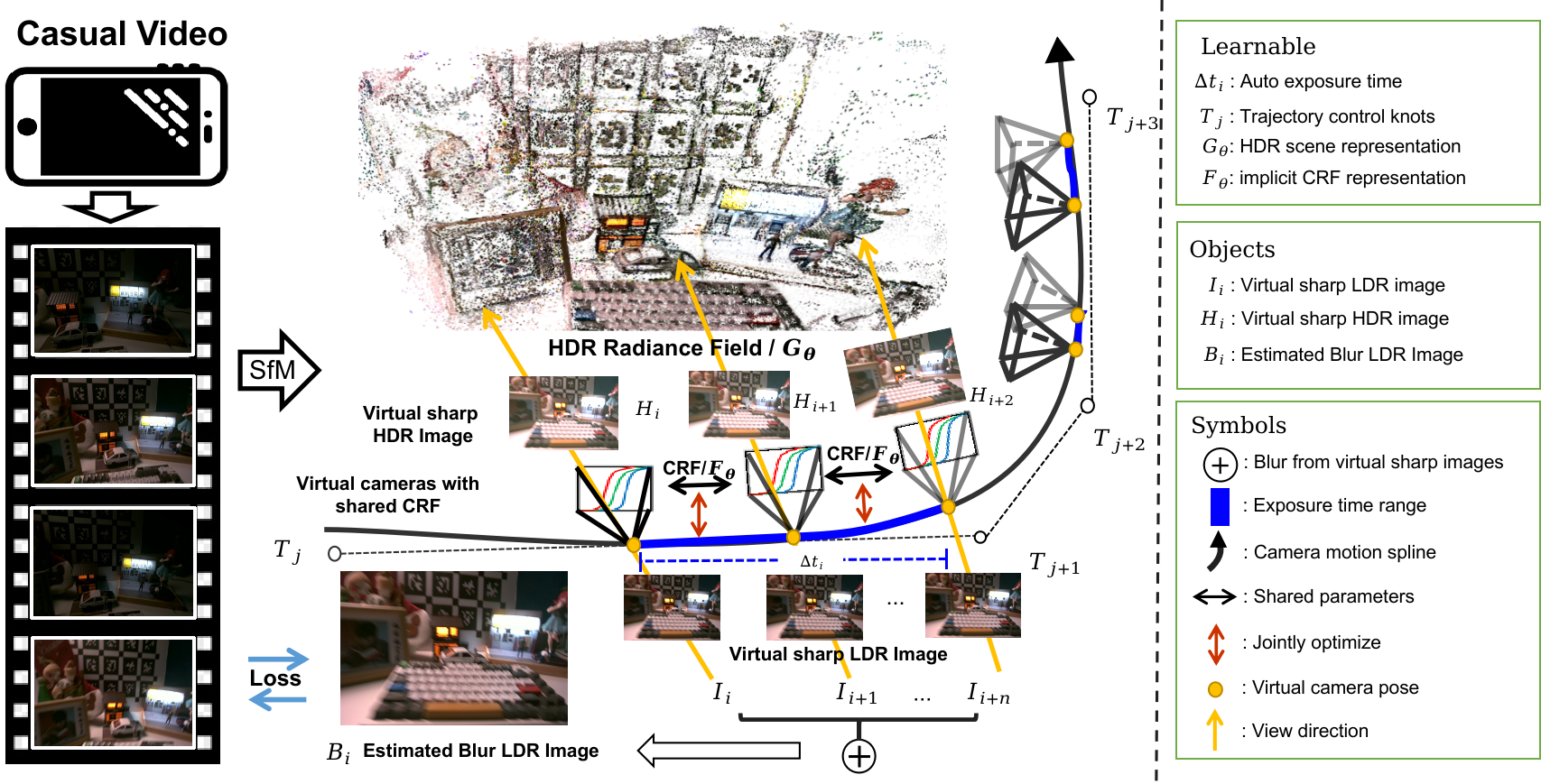}
		\vspace{-2.0em}
		\captionsetup {font={small,stretch=0.5}}

		\caption{{\bf The pipeline of Casual3DHDR.} Given a casually-captured video with AE, camera motion blur, and significant exposure time changes, we train 3DGS to reconstruct an HDR scene. We design a unified model based on the physical image formation process, integrating camera motion blur and exposure-induced brightness variations. This allows for the joint estimation of camera motion, exposure time, and camera response curve while reconstructing the HDR scene. After training, our method can sharpen the train images and render HDR images}
		\label{fig_method}
	\end{center}
\end{figure*}

\subsection{Continuous Trajectory Representation}
\label{method_traj}
Cumulative $\mathbb{SE}(3)$ B-splines are widely used in robotics for continuous-time trajectory representation, particularly in state estimation, sensor fusion, and path planning~\cite{furgale2012continuous,lovegrove2013spline,bry2015ijrr,rehder2016kalibr,mueggler2018tro,geneva2020openvins}. These splines offer desirable properties, including $C^2$ continuity, locality, and the convex hull property, which effectively incorporate gradient information and dynamic constraints, enabling rapid convergence to smooth, feasible trajectories~\cite{zhou2019robust}. Moreover, $\mathbb{SE}(3)$ B-splines allow the computation of pose, velocity, and acceleration at any timestamp along the trajectory.

Existing multi-view deblurring methods, such as BAD-NeRF~\cite{wang2023badnerf}, model camera motion by estimating short splines for each frame independently, limiting their ability to leverage cross-frame motion constraints and priors in continuous video inputs. Other approaches~\cite{Wang2021NeuralTF,Li2022DynIBaRND,sun2024_dyblurf,Li_STG_2024_CVPR,lin2024gaussian,Shih2024SiggraphAmbient,wang2024shape} employ basis functions to regularize continuous-time deformations for dynamic scene reconstruction. However, reconstructing scenes from casually-captured videos using a continuous-time camera trajectory representation remains unexplored. To address this, we propose modeling camera motion across an entire video using a cumulative $\mathbb{SE}(3)$ B-spline trajectory, enabling robust and consistent trajectory estimation.

Following \cite{lovegrove2013spline}, given a series of temporally uniformly distributed control knots, the pose $\bP(t)$ at a given timestamp $t$ can be interpolated with 4 adjacent control knots, denoted as $\bT_\mathrm{0}$, $\bT_{1}$, $\bT_{2}$ , $\bT_{3} \in \mathbb{SE}(3)$:
\begin{equation}
\bP(t) = \bT_0 \cdot \prod_{j=0}^2 \mathrm{exp}(\tilde{\bB}(u)_{j+1} \cdot \bOmega_{j}),
\end{equation}
\begin{equation}
\tilde{\bB}(u) = \bC \begin{bmatrix}
    1 \\ u \\ u^2 \\ u^3
\end{bmatrix},  \quad
\bC = \frac{1}{6} \begin{bmatrix}
    6 & 0 & 0 & 0 \\
    5 & 3 & -3 & 1 \\
    1 & 3 & 3 & -2 \\
    0 & 0 & 0 & 1
\end{bmatrix}.
\end{equation}
where $\tau$ represents the spline sampling interval, $u=\frac{t}{\tau}$, and $u$ lies within the interval $[0, 1)$; $\tilde{\bB}(u)_{j+1}$ denotes the $(j+1)^{th}$ element of the vector $\tilde{\bB}(u)$, $\bOmega_{j} = \mathrm{log}(\bT_{j}^{-1} \cdot \bT_{j+1})$, based on the \cite{Qin1998CGA}.

\begin{figure*}[h]
    \vspace{1em}
	\centering
    \includegraphics[width=\textwidth]{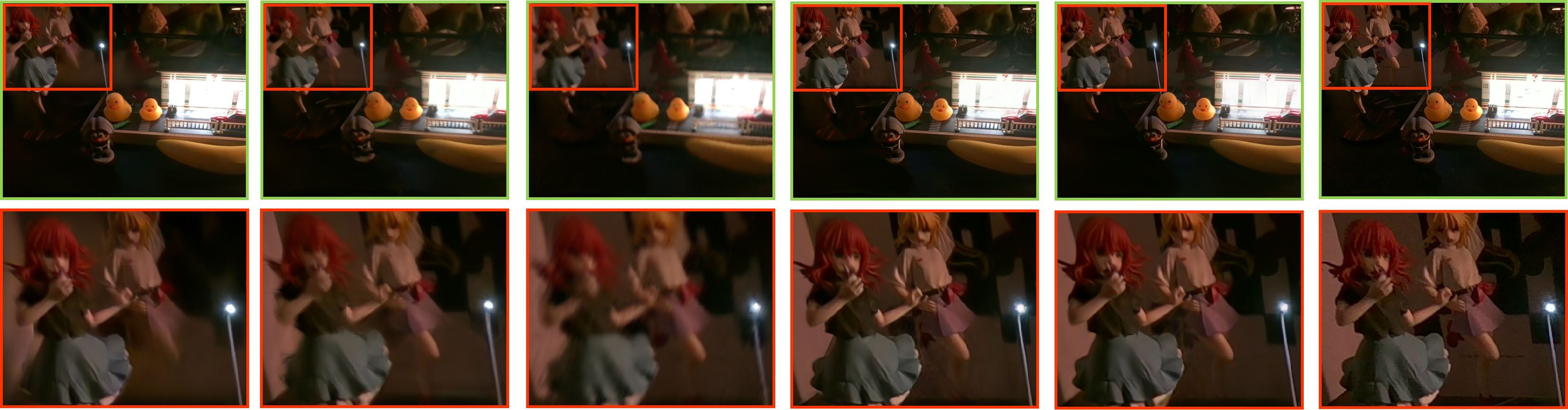}
    \setlength\tabcolsep{0pt}
	\small{
		\begin{tabular*}{\linewidth}{
			@{\extracolsep{\fill}} p{0.162\textwidth} p{0.162\textwidth} p{0.162\textwidth} p{0.162\textwidth} p{0.162\textwidth} p{0.162\textwidth}}
			BAD-Gaussians & HDR-Plenoxels & Gaussian-W & Ours-random & Ours-gt & Reference
		\end{tabular*}
	}
    \vspace{-2em}
	\caption{{\textbf{Qualitative comparison on the Girls-vicon sequence of the \textit{Realsense} dataset in terms of NVS.
 }}}
\label{girls_nvs_quali}
\end{figure*}

\begin{figure*}[h]
	\setlength\tabcolsep{1pt}
	\centering
	\includegraphics[width=\textwidth]{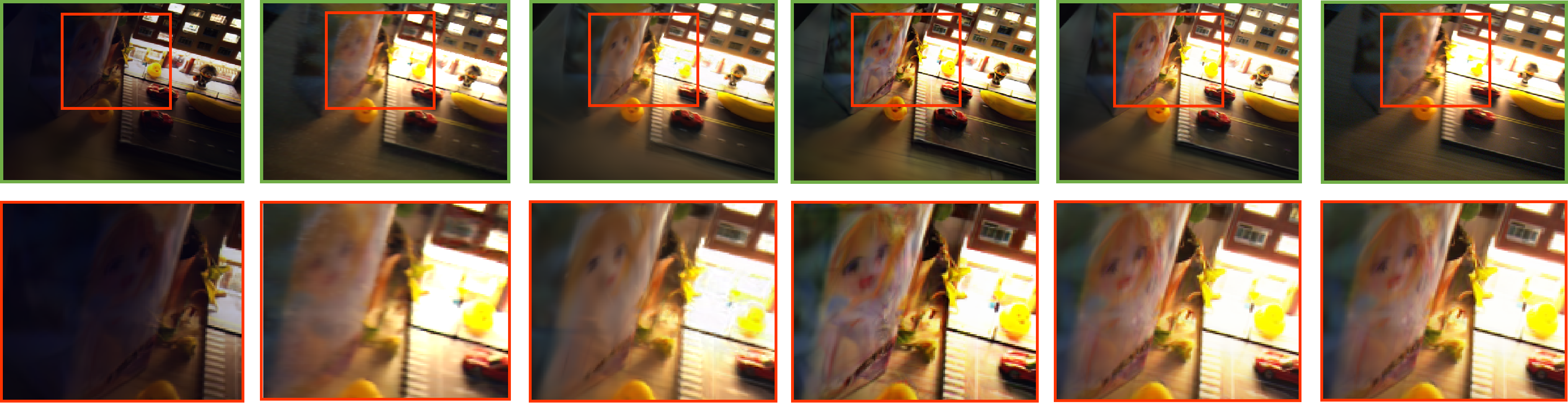}\\
	\small{
		\begin{tabular*}{
			\linewidth}{
				@{\extracolsep{\fill}}p{0.162\textwidth}p{0.162\textwidth}p{0.162\textwidth}p{0.162\textwidth}p{0.162\textwidth}p{0.162\textwidth}}
			BAD-Gaussians & HDR-Plenoxels & Gaussian-W & Ours-random & Ours-gt & Reference
		\end{tabular*}
	}
    \vspace{-2em}
	\caption{{\textbf{Qualitative comparison on the Building sequence of the \textit{Smartphone} dataset in terms of NVS.
 }}}
	\label{hotpot_nvs_quali}
\end{figure*}
\subsection{Physical Image Formation Model}
\label{method_image_model}

The physical image formation process refers to a digital camera collecting scene irradiance during the exposure time $\Delta t$ and converting them into measurable electric charges, which are ultimately mapped into pixel values through the $\textit{camera response function}$ (CRF) defined by $F$. Assuming the camera moves along a continuous trajectory $t$ $\mapsto$ $\bP(t)$ during exposure time $\Delta t$ with constant velocity, this process can be mathematically modeled as follows:

\vspace{-0.5em}
\begin{equation}\label{eq_physical_Image_formation_model}
\bB(x,y) = F \left(  \int_{t_b}^{t_b + \Delta t} \bH \left(x,y,\bP(t) \right) \mathrm{dt} \right)
\end{equation}
where $\bB(x,y) \in \nR^{\mathrm{H} \times \mathrm{W} \times 3}$ denotes the real captured image, $x,y \in \nR^2$ represents the pixel location, $t_b$ denotes the timestamp  when the shutter opens, $\bH\left(x,y,\bP(t) \right)$ represents scene irradiance mapped into camera at pose $\bP(t)$ which is interpolated from the continuous trajectory. Additionally, if the camera moves during the exposure time, the camera will collect irradiance from different scene points, resulting in camera motion blur. The integral part in Eq. (\ref{eq_physical_Image_formation_model}) can be discretized as follows:
\begin{equation}\label{eq_blur_im_formation}
	\bH(x,y)  \approx   \frac{1}{N}  \sum_{k=0}^{N-1} \bH_k \left(x,y,\bP(t_k) \right)\Delta t 
\end{equation}
 $\bH(x,y) \in \nR^{\mathrm{H} \times \mathrm{W} \times 3}$ denotes blur HDR image, N represents the number of virtual latent sharp images, $\Delta t_k$ represents the exposure time of virtual camera $k$ and can be set as a constant equal to $\frac{\Delta t}{N}$, $t_k$ denotes the timestamp corresponding to virtual camera $k$, it can be calculated as $t_b+\frac{\Delta t}{N}*k$.

After obtaining $\bH(x,y)$, we need to use the camera response function $F$, which includes image-varying white balance $\mathrm{WB}$ and tone mapping $\mathrm{TM}$, to convert it into an LDR image:
\begin{equation}
    \label{eq_hdr2ldr}
    \bB(x,y) = \bF (\bH(x,y)) = \text{TM} \circ \text{WB}(\bH(x,y))
\end{equation}
\begin{equation}
\label{eq_wb}
\text{WB}(\mathbf{c})=\begin{bmatrix}wb_r, wb_g, wb_b\end{bmatrix}^{T} \odot \begin{bmatrix}c_r, c_g, c_b\end{bmatrix}^{T}
\end{equation}
Due to the fact that RGB channels have different camera response curves for $\mathrm{TM}$, we adopt separate MLP for each channel. 
Unlike prior methods, we treat $\Delta t$ as an optimizable quantity rather than a precisely known parameter. Initially, $\Delta t$ can be assigned {\bf a random value}. Since the exposure time directly affects the brightness and motion blur of the image, it will be gradually optimized to the actual value during the subsequent deblurring and HDRI processes. This significantly reduces the dependency on the exposure time and enhances the robustness of 3D HDR reconstruction.

\subsection{Loss Function}
\label{method_loss_function}
Given a series of video frames moving along a continuous trajectory, we can estimate the learnable Gaussian primitives, the camera trajectory parameters, implicit CRF representation and the exposure time for each image. This estimation can be achieved by minimizing a loss function, which can be specifically expressed as follows:

\vspace{-1.em}
\begin{equation}
\cL = \mathcal{L}_{\text{rec}} + \lambda_{\text{exp}} \mathcal{L}_{\text{exp}},
\quad 
\mathcal{L}_{\text{rec}} = (1 - \lambda) \mathcal{L}_1 + \lambda \mathcal{L}_{\text{D-SSIM}},
\end{equation}
where $\mathcal{L}_{\text{rec}}$ can constrain the consistency between the rendered image $\bC_k(\bx)$ (the $k^{th}$ blurry LDR image synthesized from 3D-GS using \eqnref{eq_blur_im_formation}) and the input LDR image $\bC^{gt}_k(\bx)$.

To accurately model significant exposure variations in the input images, the second term of the loss function normalizes the images to a medium exposure by scaling pixel intensities before computing discrepancies \cite{liu2020single,wang2024cinematic}:
\begin{equation} 
\small
\mathcal{L}_{\text{exp}} = \mathcal{L}_1 \left( \frac{\bC^{gt}_k(\bx)}{\bar{\bC}^{gt}_k(\bx)}, \frac{\bC_k(\bx)}{\bar{\bC}_k(\bx)} \right) + \mathcal{L}_{\text{D-SSIM}} \left( \frac{\bC^{gt}_k(\bx)}{\bar{\bC}^{gt}_k(\bx)}, \frac{\bC_k(\bx)}{\bar{\bC}_k(\bx)} \right),
\end{equation}
where $\bar{\bC}^{gt}_k(\bx)$ and $\bar{\bC}_k(\bx)$ represent the average pixel value of $\bC^{gt}_k(\bx)$ and  $\bC_k(\bx)$. We set $\lambda_{\text{exp}} = 0.25$ in all our experiments.

\section{Experiments}

\begin{figure*}[t]
\setlength\tabcolsep{1pt}
\centering
  \includegraphics[width=\textwidth]{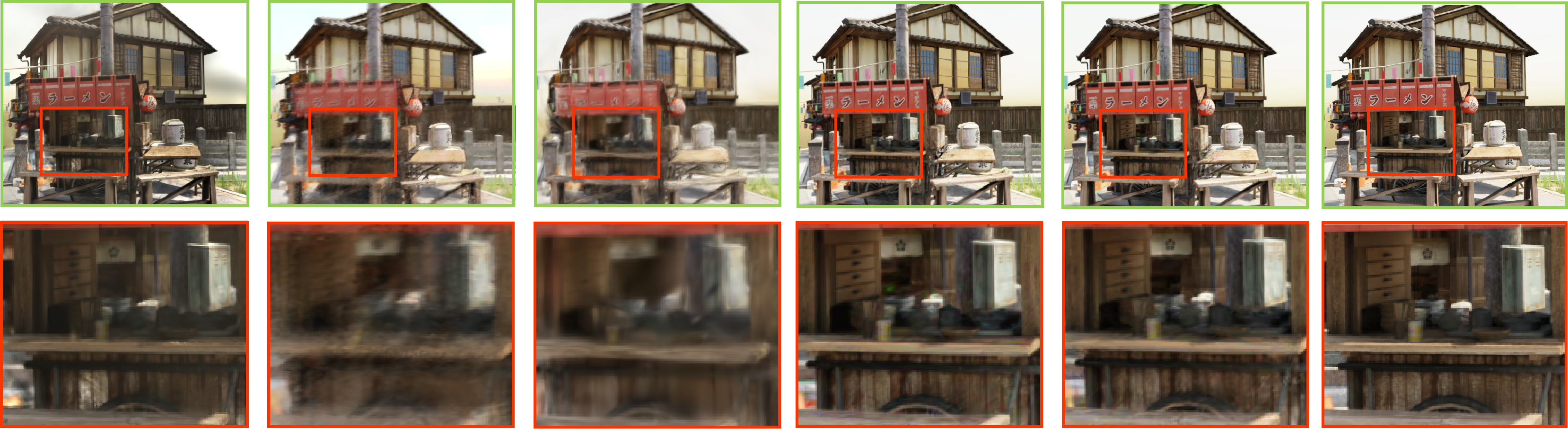}\\
  \small{
	\begin{tabular*}{
		\linewidth}{
@{\extracolsep{\fill}}p{0.162\textwidth}p{0.162\textwidth}p{0.162\textwidth}p{0.162\textwidth}p{0.162\textwidth}p{0.162\textwidth}}
		BAD-Gaussians & HDR-Plenoxel & Gaussian-W & Ours-random & Ours-gt & Reference
		\end{tabular*}
  }
\vspace{-2em}
	\caption{{\textbf{Qualitative comparison on the Trolley sequence of the \textit{synthetic} dataset in terms of NVS.
 }}}
\label{trolley_nvs_quali}
\vspace{0.5em}
\end{figure*}
\begin{figure}[!t]
\centering
\includegraphics[width=0.5\textwidth]{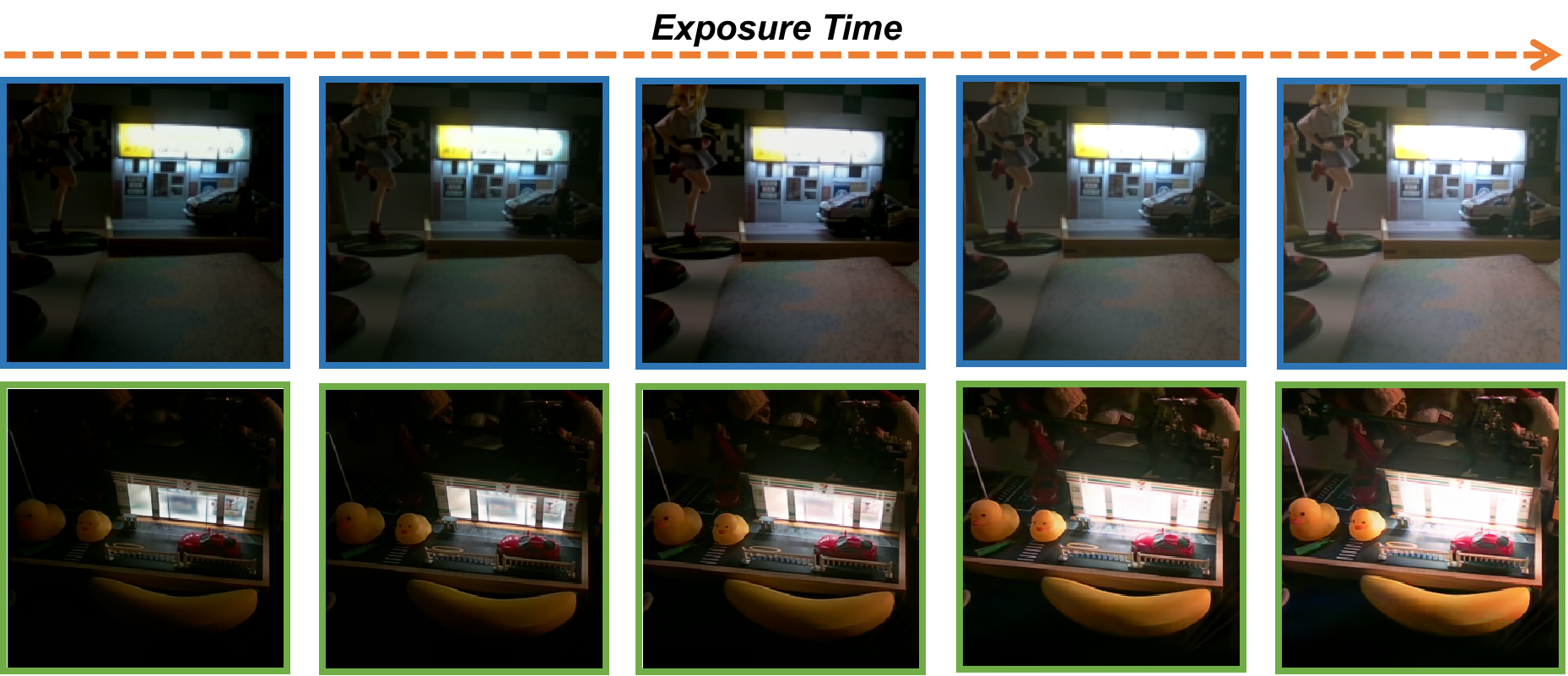}
        \vspace{-2.0em}
    	\caption{{\textbf{HDR editing with various designated exposure times.}}}
    \label{hdr_qualitative}
\end{figure}
\subsection{Datasets}
\noindent{\bf Synthetic datasets.}\quad
We created a synthetic dataset in Blender 3.6 (Cycles engine) with four photorealistic scenes: \textit{Factory}, \textit{Pool}, \textit{Cozyroom}, and \textit{Trolley}, each containing 77 images with realistic lighting and geometry. Following~\citet{wang2023badnerf, huang2022hdr}, we combined a physical imaging model and tone-mapping. Images were rendered along continuous camera trajectories with random exposure times, producing sharp HDR frames that were temporally averaged to simulate motion-blurred LDR frames.

\noindent{\bf Real datasets.}\quad
Existing HDR datasets use static-view stacks with known exposures, differing from our casual video-based setting. To bridge this gap, we collected a real-world dataset, \textit{CasualVideo}, using Intel RealSense D455 and Google Pixel 8 Pro on a DJI RS3 Mini gimbal. It includes two subsets: \textit{RealSense} (4 sequences, 2 with Vicon poses) and \textit{Smartphone} (2 sequences). As RealSense lacks exposure metadata under AE, we implemented custom AE control (fixed aperture/ISO) on both devices, following~\citet{su2015fast}. We also evaluate on ScanNet~\citep{dai2017scannet}, which uses auto-exposure (AE) ~\citep{bae2024i2slaminvertingimagingprocess}.

\begin{table*}[h]
	\begin{center}
\vspace{-0.5em}
		\captionsetup {font={small,stretch=0.5}}
		\caption{{\textbf{Quantitative comparisons on the synthetic datasets in terms of novel view}}} 
		\label{table_our_synthesis_novel_view}
		\vspace{-1em}
		\setlength\tabcolsep{2.7pt}
		\setlength{\belowcaptionskip}{-12pt}
		\scriptsize
		\resizebox{\linewidth}{!}{
			\begin{tabular}{c|ccc|ccc|ccc|ccc}
				\toprule
				&  \multicolumn{3}{c}{Factory}  &  \multicolumn{3}{|c}{Pool}  &  \multicolumn{3}{|c}{Trolley}  &  \multicolumn{3}{|c}{Cozyroom} \\
				& PSNR$\uparrow$ & SSIM$\uparrow$ & LPIPS$\downarrow$ & PSNR$\uparrow$ & SSIM$\uparrow$ & LPIPS$\downarrow$ & PSNR$\uparrow$ & SSIM$\uparrow$ & LPIPS$\downarrow$ & PSNR$\uparrow$ & SSIM$\uparrow$ & LPIPS$\downarrow$\\
				\midrule

				gsplat \citep{kerbl3Dgaussians} &15.14 & 0.75 & \cellcolor{yellow!25}0.25 
                &11.73 &0.65 & \cellcolor{yellow!25}0.30 
                &14.48 &0.62 &0.32 &13.86 &0.76 &0.22 \\

				Gaussian-W \citep{zhang2024gaussian} &23.68 & 0.75 &0.26 
                &23.28 &0.69 &0.62 
                & \cellcolor{yellow!25}17.83 &0.64 &0.34 
                &27.16 & \cellcolor{yellow!25}0.85 &0.15 \\
                
				BAD-Gaussians \citep{zhao2024badgaussians} &14.99 & \cellcolor{yellow!25}0.81 & \cellcolor{yellow!25}0.25 
                &25.09 &0.72 & \cellcolor{yellow!25}0.30 
                &16.12 & \cellcolor{yellow!25}0.66 & \cellcolor{orange!25}0.22 &17.12 &0.75 &0.20 \\
                
				HDR-Plenoxels \citep{jun2022hdr} & \cellcolor{yellow!25}24.36 &0.72 &0.29 
                & \cellcolor{yellow!25}30.84 & \cellcolor{yellow!25}0.81 &0.33 
                &17.05 &0.55 &0.42 
                & \cellcolor{yellow!25}28.13 &0.81 & \cellcolor{yellow!25}0.13 \\

				HDR-NeRF \citep{huang2022hdr} &14.57 &0.31 &0.68 &- &- &- &- &- &- &13.62 &0.32 &0.77\\

				\specialrule{0.08em}{1pt}{1pt}
    			Casual3DHDR-random (ours) &\cellcolor{orange!25}30.25 & \cellcolor{orange!25}0.89 & \cellcolor{orange!25}0.10 
                & \cellcolor{red!25}32.63 & \cellcolor{orange!25}0.91 & \cellcolor{orange!25}0.09 
                & \cellcolor{orange!25}25.14 & \cellcolor{orange!25}0.81 & \cellcolor{yellow!25}0.24 
                & \cellcolor{orange!25}29.62 & \cellcolor{red!25}0.86 & \cellcolor{red!25}0.10 \\
				
                Casual3DHDR-gt (ours) & \cellcolor{red!25}30.75 & \cellcolor{red!25}0.90 & \cellcolor{red!25}0.09 
                & \cellcolor{orange!25}32.36 & \cellcolor{red!25}0.92 & \cellcolor{red!25}0.08 & \cellcolor{red!25}25.85 & \cellcolor{red!25}0.88 & \cellcolor{red!25}0.11
                & \cellcolor{red!25}31.32 & \cellcolor{orange!25}0.92 & \cellcolor{orange!25}0.09 \\

				\specialrule{0.08em}{1pt}{1pt}
			\end{tabular}
		}
	\end{center}
\end{table*}

\begin{table*}[h]
	\begin{center}
\vspace{-0.5em}
		\captionsetup {font={small,stretch=0.5}}
		\caption{{\textbf{Quantitative comparisons on the real-world datasets in terms of novel view.}}} 
		\label{table_our_realsense_novel_view}
\vspace{-1em}
		\setlength\tabcolsep{2.7pt}
		\setlength{\belowcaptionskip}{-12pt}
		\scriptsize
		\resizebox{\linewidth}{!}{
			\begin{tabular}{c|ccc|ccc|ccc|ccc}
				\toprule
				&  \multicolumn{3}{c}{Fish-pixel8pro}  &  \multicolumn{3}{|c}{Building-pixel8pro}  &  \multicolumn{3}{|c}{Toufu-vicon}  &  \multicolumn{3}{|c}{Girls-vicon} \\
				& PSNR$\uparrow$ & SSIM$\uparrow$ & LPIPS$\downarrow$ & PSNR$\uparrow$ & SSIM$\uparrow$ & LPIPS$\downarrow$ & PSNR$\uparrow$ & SSIM$\uparrow$ & LPIPS$\downarrow$ & PSNR$\uparrow$ & SSIM$\uparrow$ & LPIPS$\downarrow$\\
				\midrule
                    gsplat &23.20 & \cellcolor{yellow!25}0.82 & 0.16 &25.99 &0.81 & \cellcolor{yellow!25}0.11 &24.34 &0.81 &0.28 &23.81 &0.77 &0.28 \\
				BAD-Gaussians &24.28 &0.78 &\cellcolor{yellow!25}0.14 &26.93 &\cellcolor{yellow!25}0.82 & \cellcolor{yellow!25}0.11 &24.22 &0.82 & \cellcolor{yellow!25}0.24 &23.95 &0.77 &0.28 \\
                    HDR-Plenoxels &19.39 &0.53 &0.65 &26.87 &0.81 &0.15 &17.90 &0.51 &0.69 &26.73 &0.84 &0.30 \\
                    Gaussian-W &\cellcolor{yellow!25}26.13 & \cellcolor{orange!25}0.83 &0.15 &\cellcolor{yellow!25}27.99 &\cellcolor{yellow!25}0.82 &\cellcolor{yellow!25}0.11 &\cellcolor{yellow!25}26.38 & \cellcolor{yellow!25}0.83 &0.29 &\cellcolor{yellow!25}26.88 & \cellcolor{yellow!25}0.86 & \cellcolor{yellow!25}0.25 \\
				\specialrule{0.08em}{1pt}{1pt}
                Casual3DHDR-random (ours) &\cellcolor{orange!25}28.30 & \cellcolor{orange!25}0.83 & \cellcolor{orange!25}0.13 &\cellcolor{orange!25}28.79 & \cellcolor{orange!25}0.83 &\cellcolor{orange!25}0.09 &\cellcolor{orange!25}30.87 & \cellcolor{orange!25}0.90 &\cellcolor{orange!25}0.15 &\cellcolor{orange!25}32.00 & \cellcolor{orange!25}0.90 & \cellcolor{orange!25}0.19 \\
				Casual3DHDR-gt (ours) &\cellcolor{red!25}30.81 & \cellcolor{red!25}0.87 & \cellcolor{red!25}0.12 &\cellcolor{red!25}29.71 & \cellcolor{red!25}0.85 &\cellcolor{red!25}0.08 &\cellcolor{red!25}31.34 & \cellcolor{red!25}0.92 &\cellcolor{red!25}0.12 &\cellcolor{red!25}32.39 & \cellcolor{red!25}0.91 & \cellcolor{red!25}0.17 \\
				\specialrule{0.08em}{1pt}{1pt}
			\end{tabular}
		}
	\end{center}
\end{table*}

\subsection{Implementation Details}
We implemented our method in PyTorch using the \texttt{gsplat}~\citep{ye2024gsplatopensourcelibrarygaussian} with MCMC strategy~\citep{kheradmand20243d}. The optimization of all parameters was done using the Adam optimizer, keeping the Gaussian primitive learning rate consistent with \texttt{gsplat}. To balance performance and efficiency, we set the number of virtual camera poses (n in \eqnref{eq_blur_im_formation}) to 10.

For initialization, we used HLoc~\citep{sarlin2019coarse} on our \textit{Sythetic} and \textit{Realsense} dataset. For the ScanNet~\citep{dai2017scannet} and our \textit{Smartphone} datasets, we used DPV-SLAM~\citep{lipson2024deep} since HLoc failed due to poor image quality. Synthetic dataset experiments were conducted on an NVIDIA RTX 3090, while real dataset experiments used an NVIDIA RTX A6000 due to higher memory requirements.

\subsection{Baseline Methods and evaluation metrics}
To evaluate the robustness of our method in learning accurate scenes representation under poorly exposed conditions and server motion blur, we compared it against scene reconstruction methods that handle brightness variations, e.g. HDR-NeRF~\citep{huang2022hdr}, HDR-Plenoxels~\citep{jun2022hdr}, Gaussian-W~\citep{zhang2024gaussian}, as well as method for scene reconstruction from blurred images, such as BAD-Gaussians~\citep{zhao2024badgaussians}. In addition, 3D-GS~\citep{kerbl3Dgaussians} implemented by gsplat~\citep{ye2024gsplatopensourcelibrarygaussian} was included as the comparison baseline. Following previous works~\cite{jun2022hdr,cai2024hdrgsefficienthighdynamic,huang2022hdr}, we evaluate LDR images rendered from the learned scene used metrics such as PSNR, SSIM~\citep{zhou2004tipssim}, and LPIPS~\citep{zhang2018unreasonableeffectivenessdeepfeatures}. Furthermore, to evaluate whether our method effectively recovers camera motion trajectories, we compared it with pose estimation method, e.g. HLoc~\citep{sarlin2019coarse}, DPV-SLAM~\citep{lipson2024deep}, BAD-Gaussians~\citep{zhao2024badgaussians}. For pose estimation, we utilize absolute trajectory error (ATE) with \textit{mean} and \textit{std} as the metric.

\subsection{Quantitative evaluation results.}
We conducted experiments with two settings: one using randomly initialized exposure times (\textbf{Casual3DHDR-random}) and the other with ground truth exposure times (\textbf{Casual3DHDR-gt}). We evaluated our method's performance on novel view synthesis, image deblurring, and pose estimation tasks. More results on the ScanNet and \textit{Realsense} scenes can be found in the supplementary material.

Due to the blurry nature of most real-world images, we used 5 to 10 sharp images per sequence for evaluation. As shown in \tabnref{table_our_synthesis_novel_view} and \tabnref{table_our_realsense_novel_view}, our method significantly outperforms previous methods in novel view synthesis. \textbf{Casual3DHDR-random}, even with randomly initialized exposure times, surpasses prior works by jointly optimizing exposure times and CRF representation. Unlike HDR-NeRF~\citep{huang2022hdr}, our method reconstructs HDR scenes accurately from degraded images without measured exposure times, whereas HDR-NeRF fails on all real dataset scenes.

By modeling physical principles of imaging, our method also outperforms HDR-Plenoxels~\citep{jun2022hdr} and Gaussian-W~\citep{zhang2024gaussian}. Moreover, our approach leverages spline representations for optimizing camera motion trajectories, enabling effective scene learning, which other methods struggle with the absence of ground truth poses.

\begin{table*}[h]
	\begin{center}
\vspace{-0.5em}
		\captionsetup {font={small,stretch=0.5}}
		\caption{{\textbf{Quantitative comparisons on the synthetic datasets in terms of deblur}}} 
		\label{table_our_synthesis_deblur}
\vspace{-1em}
		\setlength\tabcolsep{2.7pt}
		\setlength{\belowcaptionskip}{-12pt}
		\scriptsize
		\resizebox{\linewidth}{!}{
			\begin{tabular}{c|ccc|ccc|ccc|ccc}
				\toprule
				&  \multicolumn{3}{c}{Factory}  &  \multicolumn{3}{|c}{Pool}  &  \multicolumn{3}{|c}{Trolley}  &  \multicolumn{3}{|c}{Cozyroom} \\
				& PSNR$\uparrow$ & SSIM$\uparrow$ & LPIPS$\downarrow$ & PSNR$\uparrow$ & SSIM$\uparrow$ & LPIPS$\downarrow$ & PSNR$\uparrow$ & SSIM$\uparrow$ & LPIPS$\downarrow$ & PSNR$\uparrow$ & SSIM$\uparrow$ & LPIPS$\downarrow$\\
				\midrule

				BAD-Gaussians~\citep{zhao2024badgaussians} 
                & \cellcolor{yellow!25}24.32 & \cellcolor{yellow!25}0.73 & \cellcolor{yellow!25}0.12 
                & \cellcolor{yellow!25}25.87 & \cellcolor{yellow!25}0.79 & \cellcolor{yellow!25}0.23 
                & \cellcolor{yellow!25}19.06 & \cellcolor{yellow!25}0.62 & \cellcolor{yellow!25}0.19 
                & \cellcolor{yellow!25}23.37 & \cellcolor{yellow!25}0.79 & \cellcolor{yellow!25}0.11 \\

				\specialrule{0.08em}{1pt}{1pt}
                Casual3DHDR-random (ours) 
                & \cellcolor{orange!25}31.20 & \cellcolor{orange!25}0.88 & \cellcolor{red!25}0.05 
                & \cellcolor{orange!25}32.95 & \cellcolor{orange!25}0.87  & \cellcolor{orange!25}0.10 
                & \cellcolor{orange!25}23.65 & \cellcolor{orange!25}0.69 & \cellcolor{orange!25}0.12 
                & \cellcolor{orange!25}29.60 & \cellcolor{orange!25}0.84 & \cellcolor{orange!25}0.05   \\
                
				Casual3DHDR-gt (ours) 
                & \cellcolor{red!25}32.00 & \cellcolor{red!25}0.91 & \cellcolor{orange!25}0.07 
                & \cellcolor{red!25}34.53 & \cellcolor{red!25}0.96 & \cellcolor{red!25}0.05 
                & \cellcolor{red!25}29.35 & \cellcolor{red!25}0.87 & \cellcolor{red!25}0.08 
                & \cellcolor{red!25}33.01 & \cellcolor{red!25}0.93 & \cellcolor{red!25}0.04   \\

				\specialrule{0.08em}{1pt}{1pt}
			\end{tabular}
		}
	\end{center}
\end{table*}

\begin{table*}[h]
    \begin{center}
        \vspace{-0.5em}
        \captionsetup {font={small,stretch=0.8}}
        \caption{{\textbf{Quantitative comparisons on the ScanNet dataset in term of deblur (using BRISQUE metric).}}}
        \vspace{-1em}
        \label{table_brisque_comparison}
        \setlength\tabcolsep{2pt}
        \setlength{\belowcaptionskip}{-10pt}
        \scriptsize
        \resizebox{\linewidth}{!}{
            \begin{tabular}{l|cccccc|c}
                \toprule
                & \textbf{scene0024\_01} & 
                \textbf{scene0031\_00} & 
                \textbf{scene0036\_00} & \textbf{scene0072\_01} & \textbf{scene0077\_00} & \textbf{scene0489\_02} &  \textbf{Average} \\
                \midrule
                BAD-Gaussians \citep{zhao2024badgaussians} & 
                \cellcolor{orange!25}42.78 & 
                \cellcolor{orange!25}60.48 & 
                \cellcolor{orange!25}57.37 & \cellcolor{orange!25}42.72 & \cellcolor{orange!25}67.30 & \cellcolor{orange!25}65.01 & \cellcolor{orange!25} 55.94 \\
                Casual3DHDR-random (ours) & 
                \cellcolor{red!25}38.08 & 
                \cellcolor{red!25}47.65 & 
                \cellcolor{red!25}53.37 & \cellcolor{red!25}37.55 & \cellcolor{red!25}62.35 & \cellcolor{red!25}58.23 & \cellcolor{red!25} 49.53 \\
                \bottomrule
            \end{tabular}
        }
    \end{center}
\end{table*}

\begin{table*}[!h]
    \begin{center}
        \captionsetup {font={small,stretch=0.5}}
\vspace{-0.5em}
        \caption{\textbf{Quantitative comparisons for pose estimation on the \textit{Realsense} sequences with Vicon motion captured groundtruth.} The results are in the absolute trajectory error metric (ATE) with units in centimeters.}
\vspace{-1em}
        \label{table_ate_on_realsense}

        \normalsize %
        \setlength{\tabcolsep}{4mm} %
        \setlength{\belowcaptionskip}{-10pt}
        \begin{tabular*}{\linewidth}{@{\extracolsep{\fill}}c|ccccc}
            \toprule
            \makebox[0.08\textwidth][c]{} & \makebox[0.08\textwidth][c]{\normalsize HLoc~\citep{sarlin2019coarse}} & \makebox[0.08\textwidth][c]{\normalsize DPV-SLAM~\citep{lipson2024deep}} & \makebox[0.08\textwidth][c]{\normalsize BAD-Gaussians~\citep{zhao2024badgaussians}} &
            \makebox[0.08\textwidth][c]{\normalsize Casual3DHDR-random} & \makebox[0.08\textwidth][c]{\normalsize Casual3DHDR-gt} \\
            \midrule
            Toufu-vicon        &{.4644}$\pm${.3921} &{.4043}$\pm${.3877} &\cellcolor{yellow!25}{.3935}$\pm${.4212} &
            \cellcolor{orange!25}{.3687}$\pm${.3874} &\cellcolor{red!25}{.3595}$\pm${.3462} \\
            Girls-vicon    &{1.528}$\pm${1.011}    
            &{.9557}$\pm${.8231}  &\cellcolor{yellow!25}{.8548}$\pm${.8628} &\cellcolor{orange!25}{.8294}$\pm${.8834} &\cellcolor{red!25}{.6478}$\pm${.8268} \\
            \bottomrule
        \end{tabular*}
    \end{center}
\end{table*}

\begin{table*}[!h]
    \centering
\vspace{-0.5em}
    \begin{minipage}{0.47\textwidth} %
        \centering
        \caption{{\bf{Ablation studies on the interpolation $ratio$ for initializing camera motion spline.}}}
        \label{abation_on_ratio}
\vspace{-1.0em}
        \setlength\tabcolsep{6pt}
        \resizebox{1.0\columnwidth}{!}{
            \renewcommand{\arraystretch}{1.0}
            \begin{small}

            \begin{tabular}{c|ccc|ccc}
                \specialrule{0.1em}{1pt}{1pt}
                & \multicolumn{3}{c|}{Pool} & \multicolumn{3}{c}{Factory} \\
                $ratio$ & {\scriptsize PSNR$\uparrow$} & {\scriptsize SSIM$\uparrow$} & {\scriptsize LPIPS$\downarrow$} & {\scriptsize PSNR$\uparrow$} & {\scriptsize SSIM$\uparrow$} & {\scriptsize LPIPS$\downarrow$} \\
                \specialrule{0.1em}{1pt}{1pt}
                0.5 & 29.89 & 0.83 & 0.12 & 23.25 & 0.69 & 0.20 \\
                1.0 & 30.49 & 0.83 & 0.11 & 23.93 & 0.70 & \cellcolor{yellow!25}0.16 \\
                1.5 & 31.13 & 0.84 & \cellcolor{yellow!25}0.10 & 24.78 & 0.74 & \cellcolor{yellow!25}0.16 \\
                2.0 & 32.01 & 0.87 & \cellcolor{yellow!25}0.10 & 25.64 & \cellcolor{yellow!25}0.77 & \cellcolor{yellow!25}0.16 \\
                2.5 & 32.04 & \cellcolor{yellow!25}0.88 & \cellcolor{yellow!25}0.10 & 26.90 & \cellcolor{orange!25}0.81 & \cellcolor{red!25}0.14 \\
                3.0 & \cellcolor{yellow!25}32.95 & \cellcolor{yellow!25}0.88 & \cellcolor{yellow!25}0.10 & \cellcolor{yellow!25}27.25 & \cellcolor{red!25}0.84 & \cellcolor{orange!25}0.15 \\
                3.5 & \cellcolor{orange!25}33.13 & \cellcolor{orange!25}0.89 & \cellcolor{orange!25}0.09 & \cellcolor{orange!25}27.54 & \cellcolor{red!25}0.84 & \cellcolor{red!25}0.14 \\
                4.0 & \cellcolor{red!25}33.63 & \cellcolor{red!25}0.90 & \cellcolor{red!25}0.08 & \cellcolor{red!25}27.60 & \cellcolor{red!25}0.84 & \cellcolor{red!25}0.14 \\
                \specialrule{0.1em}{1pt}{1pt}
            \end{tabular}
            \end{small}
        }
    \end{minipage}%
    \hspace{0.05\textwidth} %
    \begin{minipage}{0.47\textwidth} %
        \centering
        \caption{\textbf{Ablation study: Average results across two datasets to investigate the effect on model performance.}}
        \label{ablation_on_module_avg}
        \vspace{-1.0em}
        \setlength\tabcolsep{4pt}
        \resizebox{1.0\columnwidth}{!}{
            \renewcommand{\arraystretch}{1.2}
            \begin{small}
            \begin{tabular}{cccc|ccc}
                \specialrule{0.1em}{1pt}{1pt}
                \multirow{2}{*}{Deblur} &
                \multirow{2}{*}{\begin{tabular}[c]{@{}c@{}}Exp. \\ Opt.\end{tabular}} &
                \multirow{2}{*}{CRF} &
                \multirow{2}{*}{\begin{tabular}[c]{@{}c@{}}Conti. \\ Traj.\end{tabular}} &
                \multicolumn{3}{c}{Average} \\
                &  &  &  & PSNR$\uparrow$ & SSIM$\uparrow$ & LPIPS$\downarrow$ \\
                \specialrule{0.1em}{1pt}{1pt}
                \xmark & \xmark & \xmark & \xmark & 14.50 & 0.755 & 0.235 \\
                \cmark & \xmark & \xmark & \xmark & 16.06 & 0.78 & 0.225 \\
                \xmark & \xmark & \xmark & \cmark & 20.04 & 0.665 & 0.29 \\
                \cmark & \xmark & \xmark & \cmark & 19.98 & 0.675 & 0.29 \\
                \xmark & \cmark & \cmark & \cmark & 26.03 & 0.79 & 0.12 \\
                \cmark & \cmark & \cmark & \cmark & 28.43 & 0.84 & 0.10 \\
                \specialrule{0.1em}{1pt}{1pt}
            \end{tabular}
            \end{small}
        }
    \end{minipage}
\vspace{-2em}
\end{table*}

\begin{figure*}[h]
\vspace{2em}
\includegraphics[width=\textwidth]{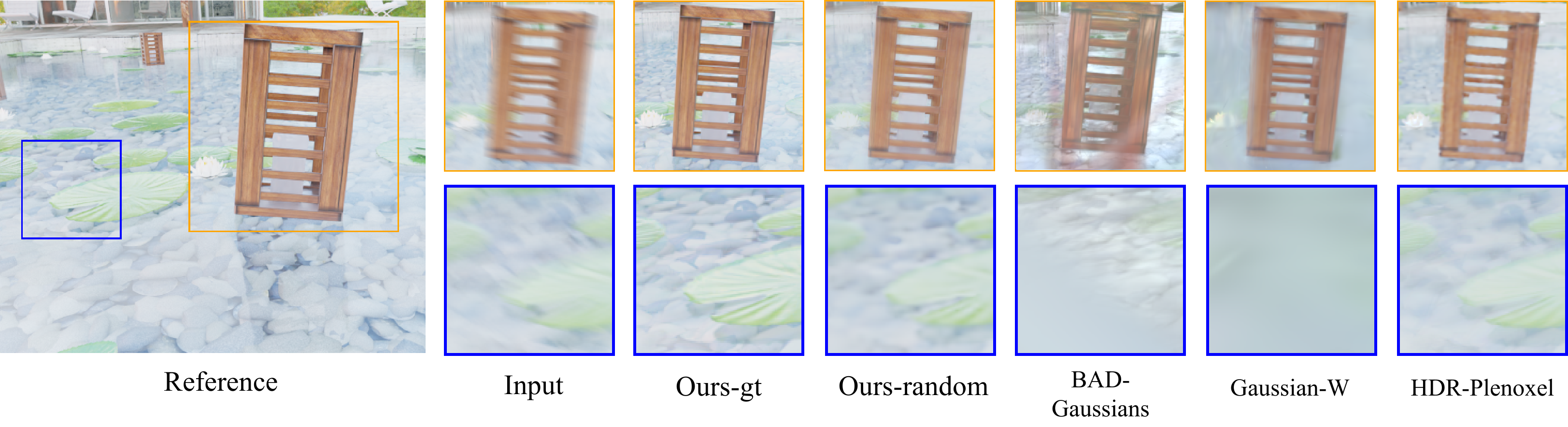}
\vspace{-2em}
\caption{\textbf{Qualitative comparison on the Pool sequence of the \textit{synthetic} dataset under training view. BAD-Gaussians is capable to
deblur the training views as ours
 } Due to the failure of pose optimization in the BAD-Gaussians, its image are misaligned with others. }
\label{pool_deblur_quali}
\end{figure*}

\tabnref{table_our_synthesis_deblur} shows that our method achieves superior performance in image deblurring task compared to BAD-Gaussians~\citep{zhao2024badgaussians}. This is because our method can recover accurate scene representation from images affected by both motion blur and poor exposure.

In addition, we also evaluated the quantitative metrics for deblurring on public real-world datasets (i.e. ScanNet \citep{dai2017scannet}). Due to most images of ScanNet dataset are motion-blurred, we can not find sharp reference images for evaluating, thus we utilize the no-reference image quality metric BRISQUE \citep{brisque} to quantitatively compare the deblurring performance between our method and BAD-Gaussians \citep{zhao2024badgaussians}, as shown in Table \ref{table_brisque_comparison}.

The experimental results presented in \tabnref{table_ate_on_realsense} demonstrate that our method outperforms prior approaches in the pose estimation task. HLoc, which relies on feature point matching, exhibits poor performance under conditions of varying brightness and motion blur. Although DPV-SLAM~\citep{lipson2024deep} and BAD-Gaussians~\citep{zhao2024badgaussians} can operate effectively in the presence of motion blur, they struggle to tolerate environments with high-contrast and varying exposure time. This indicates that our method can robustly estimate continuous camera trajectories under high-contrast environments, within varying exposure time and motion blur.

\subsection{Qualitative evaluation results.}
The results in \figrefer{hdr_qualitative} demonstrate that our method can accurately recover HDR scenes, while the brightness of the rendered images can be manually adjusted by changing the exposure time.
Qualitative comparisons of the NVS and deblurring tasks on both synthetic and real datasets are shown in \figrefer{girls_nvs_quali}, \figrefer{hotpot_nvs_quali}, \figrefer{trolley_nvs_quali} and \figrefer{pool_deblur_quali}. The experimental results indicate that our method outperforms previous approaches and is visually closer to the ground truth. This demonstrates that our method can effectively recover high-quality 3D HDR scene from images that simultaneously exhibit varying exposure time and motion blur, while prior work lacks robustness given the challenging conditions.

\subsection{Ablation studies.}

\noindent{\bf Initialization for camera motion spline.}\quad
 In our method, the camera motion spline needs to be initialized by leveraging the poses estimated from HLoc \citep{sarlin2019coarse} or DPV-SLAM \citep{lipson2024deep}. Therefore, the configuration of initialization will impact the performance of our method. We define a \textit{ratio} of interpolation representing the number of control knots of spline divided by the number of input images and evaluate the effect of the \textit{ratio}. The results in \tabnref{abation_on_ratio} indicate that model performance improves until it saturates as the \textit{ratio} increases. We set $\textit{ratio}=3.0$ for all experiments to ensure a trade-off between the performance and computational overhead.

\noindent{\bf Effect of each module.}\quad
Deblur represents the method's ability to remove blur, Exp. Opt. indicates exposure time optimization, CRF represents whether the model includes a CRF module, and Conti. Traj. refers to the use of continuous trajectories to represent camera motion. The results presented in \tabnref{ablation_on_module_avg} highlight several key findings: 1) Utilizing splines to represent the continuous camera trajectory significantly enhances model performance, achieving approximately a 24\% improvement in PSNR. 2) Jointly optimizing exposure time while learning an implicit representation of the CRF substantially boosts performance, leading to a 42\% increase in PSNR. This demonstrates that our method can robustly reconstruct HDR scenes in environments with varying brightness. 3) Representing motion blur as the average of a series of sharp images over the exposure time yields a 9\% improvement in PSNR, showing that our approach effectively handles input images with motion blur.
In summary, the proposed representation of continuous trajectories and the joint optimization of exposure time with CRF contribute significantly to the model's performance.

\vspace{-0.5em}
\section{Conclusion}

In this paper, we present Casual3DHDR, a novel framework for reconstructing high-quality 3D HDR scenes from casually-captured auto-exposure (AE) videos, which often exhibit limited dynamic range, motion blur, and uncontrolled camera motion. Our approach effectively reconstructs 3D HDR scenes and generates photorealistic LDR images for specified exposures and camera poses, demonstrating robustness and versatility. By integrating latent motion and temporal dynamic range from AE videos into a unified physical imaging model, our method jointly optimizes exposure time, continuous camera trajectories, and the camera response function (CRF). Extensive experiments validate the accuracy and effectiveness of Casual3DHDR, highlighting its potential to advance 3D HDR scene reconstruction from challenging real-world inputs.

\newcolumntype{L}[1]{>{\raggedright\let\newline\\\arraybackslash\hspace{0pt}}m{#1}}
\newcolumntype{C}[1]{>{\centering\let\newline\\\arraybackslash\hspace{0pt}}m{#1}}
\newcolumntype{R}[1]{>{\raggedleft\let\newline\\\arraybackslash\hspace{0pt}}m{#1}}

\section{Acknowldgement}
This work was supported in part by the National Natural Science Foundation of China (Grants 62202389, 42371451, 42394061), in part by the Natural Science Foundation of Wuhan (No.2024040701010028),
in part by a grant from the Westlake University-Muyuan Joint Research Institute and in part by the Westlake Education Foundation. We specially thank Xiang Liu for his kind and valuable suggestions during the writing of this paper.

\clearpage
\appendix

\section{Appendix}
\label{sec:rationale}
In the supplementary material, we present more quantitative and qualitative experimental results for image rendering under both training and novel viewpoints. We also visualized the results of camera motion estimation and performed a qualitative comparison. The rendered novel view high frame-rate HDR video is presented in the supplementary video. We will present each part as follows. 



\subsection{More experimental results of exposure time estimation.}

\begin{figure*}[h]
    \includegraphics[width=0.45\linewidth]{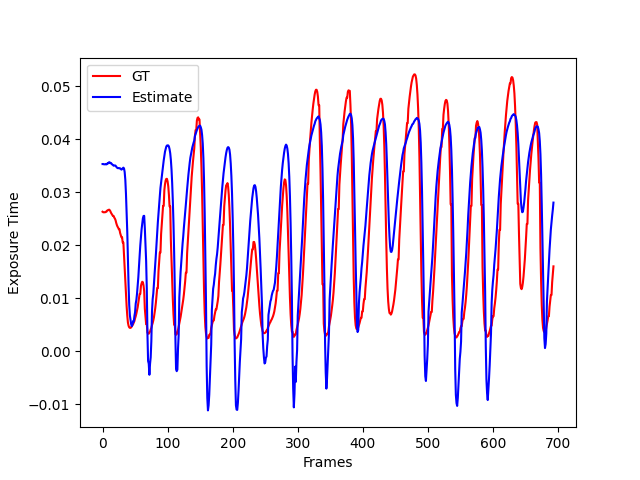}
    \includegraphics[width=0.45\linewidth]{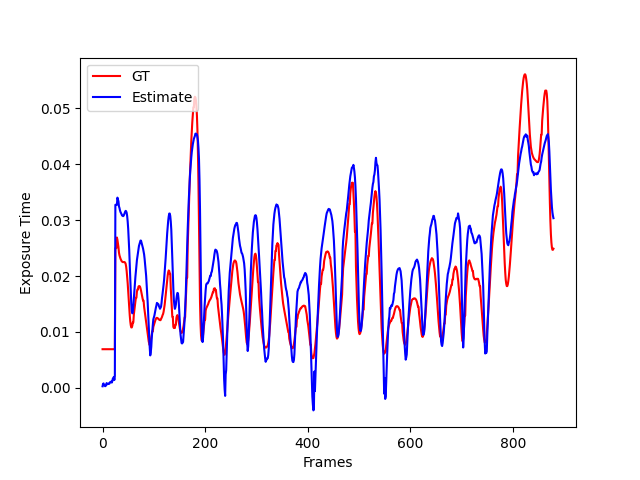}
    \vspace{-1.8em}
    \caption{Comparison between estimated exposure times and ground truth exposure times. Girls-vicon (left) and Toufu-vicon (right).}
    \label{exposure_time_estimation}
\end{figure*}

\begin{table}[h]
	\begin{center}
		\vspace{1em}
		\captionsetup {font={small,stretch=0.5}}
		\caption{{\textbf{Quantitative results for exposure time estimation.}}} 
		\label{table_exposure_time_estimation}
		\vspace{-0.5em}
		\setlength\tabcolsep{12pt}
		\setlength{\belowcaptionskip}{-12pt}
		\small
		\resizebox{\linewidth}{!}{
			\begin{tabular}{lccc}
				\toprule
				\textbf{Scene} & \textbf{Pearson} & \textbf{Spearman} & \textbf{Kendall} \\
				\midrule
				Toufu-vicon & 0.799 & 0.871 & 0.871 \\
				Girls-vicon & 0.843 & 0.865 & 0.865 \\
				\bottomrule
			\end{tabular}
		}
	\end{center}
	\vspace{-1em}

\end{table}
In physical imaging formation process, the camera first captures a RAW image $Y(p)$, and the relationship between the RAW image $Y(p)$ and the scene radiance $X(p)$ is as follows:
\[
X(p) = \frac{Y(p)}{t g \pi \left( \frac{f}{2 a} \right)^2} 
\]
where $f$ is the focal length, $t$ is the exposure time, $g$ is the gain, $a$ is the aperture f-number. Like most 3D HDR methods\cite{jun2022hdr,cai2024hdrgsefficienthighdynamic,huang2022hdr}, the $f$, $g$ and $a$ are fixed but unknown in experiments. Thus, the exposure times that we estimate have a constant scale with the ground truth exposure times. To evaluate the exposure time estimation, we used metrics such as Pearson correlation coefficient, Spearman rank correlation coefficient, and Kendall Tau-b correlation coefficient.  A correlation value greater than 0.7 is generally considered a strong correlation. The quantitative results are shown in \tabnref{table_exposure_time_estimation}, the results show that we can estimate exposure times :

\figrefer{exposure_time_estimation} shows the comparison between the jointly-optimized exposure times and the ground truth exposure times for each training image in scenes Toufu-vicon (left) and Girls-vicon (right). The results are scaled uniformly, and it can be observed that the estimated exposure times closely follow the trend of the ground truth exposure times.

\subsection{More experimental results under training view.}
\label{sec:appendix_experiment}


\begin{figure*}[h]
\includegraphics[width=.95\textwidth]{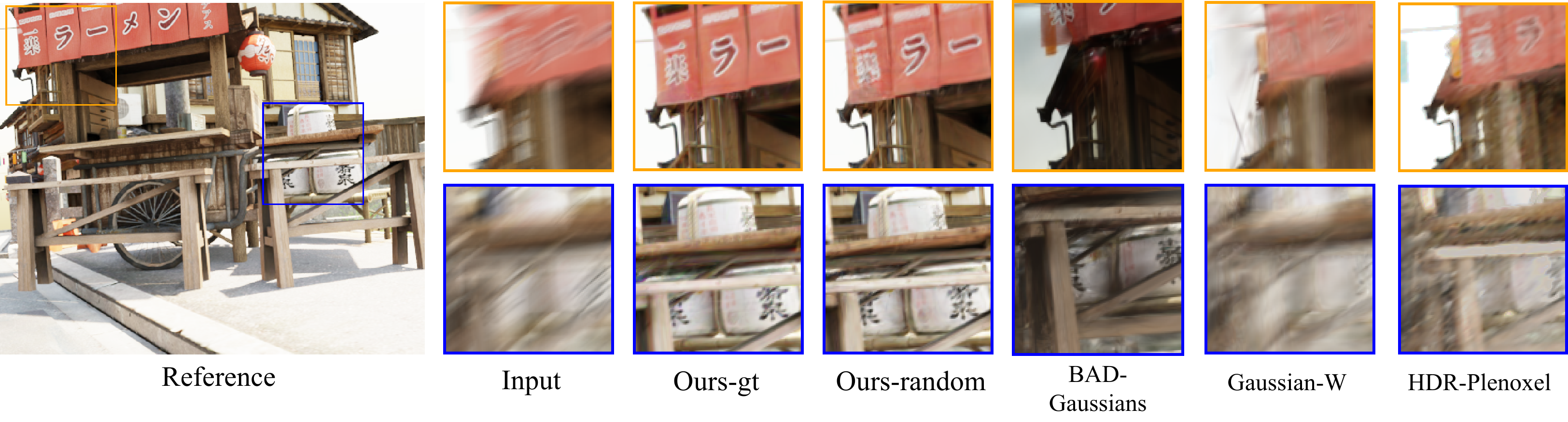}
\includegraphics[width=.95\textwidth]
{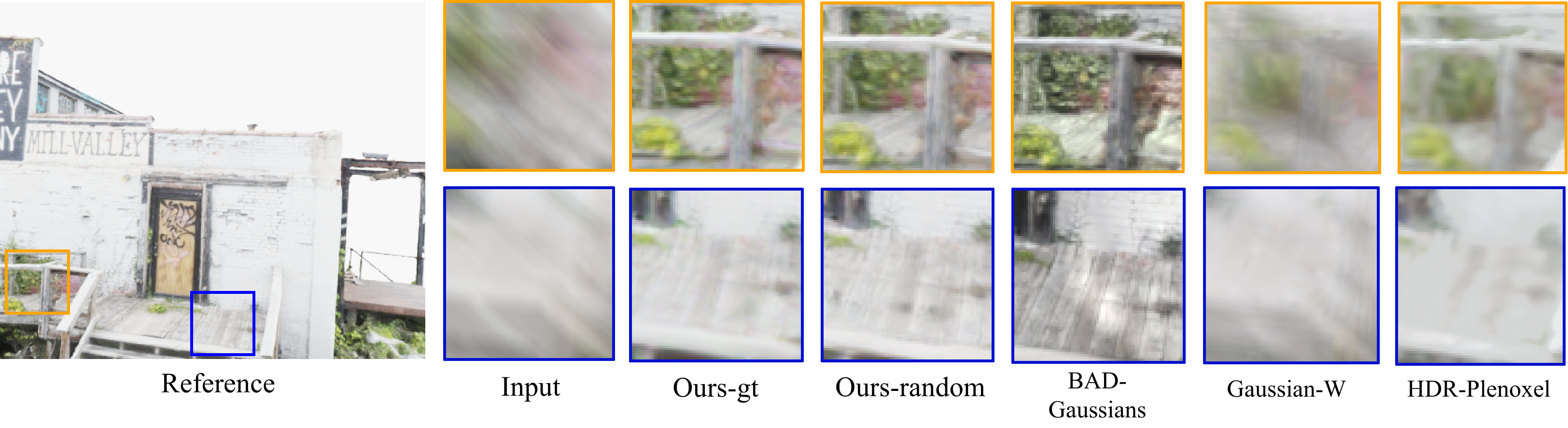}
\includegraphics[width=.95\textwidth]
{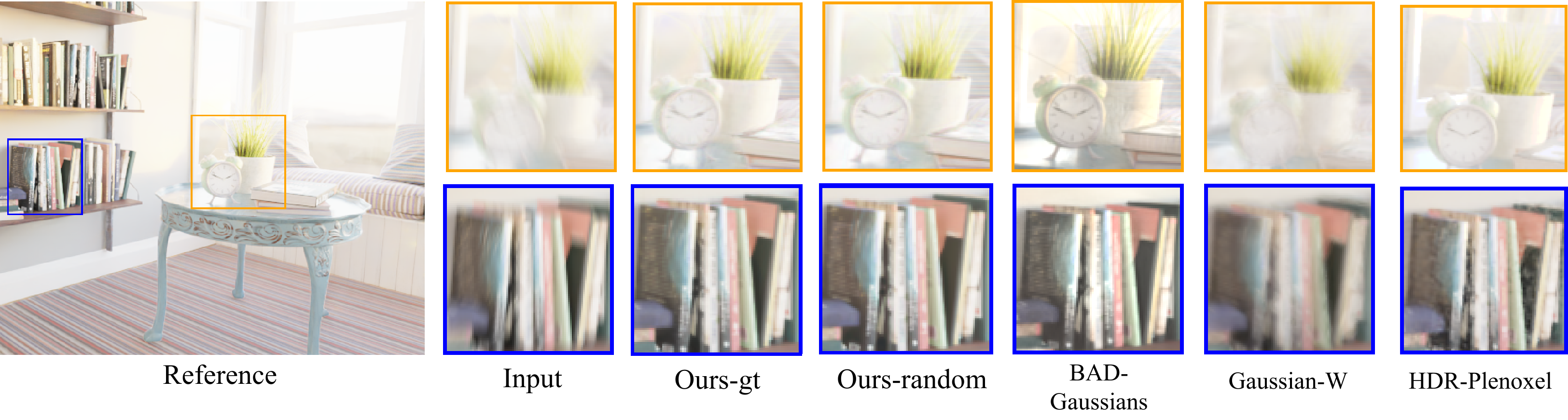}
\vspace{-1em}
\caption{\textbf{Qualitative comparison on \textit{synthetic} dataset (Trolley, Factory, Cozyroom) under training view.} BAD-Gaussians can deblur the training views as ours. However, due to the failure of pose optimization in the BAD-Gaussians, its image are misaligned with others.}
	\label{more_deblur_quali_synthetic}
\end{figure*}

The results in \figrefer{more_deblur_quali_synthetic} demonstrate that our method effectively deblurs images under training views and achieves better image quality compared to other methods. It is worth noting that while BAD-Gaussians \citep{zhao2024badgaussians} is also capable of deblurring images under training view, its lack of robustness to varying brightness conditions leads to pose optimization failure. As a result, The performance of deblurring is poor, even causing misalignment in the images under the training view.

\begin{table*}[h]
	\begin{center}
		\vspace{1em}
		\captionsetup {font={small,stretch=0.5}}
		\caption{{\textbf{Quantitative comparisons on the synthetic datasets in term of deblur.}}} 
		\label{more_table_our_synthesis_deblur}
		\vspace{-1em}
		\setlength\tabcolsep{2.7pt}
		\setlength{\belowcaptionskip}{-12pt}
		\scriptsize
		\resizebox{\linewidth}{!}{
			\begin{tabular}{l|ccc|ccc|ccc|ccc}
				\toprule
				&  \multicolumn{3}{c}{Factory}  &  \multicolumn{3}{|c}{Pool}  &  \multicolumn{3}{|c}{Trolley}  &  \multicolumn{3}{|c}{Cozyroom} \\
				& PSNR$\uparrow$ & SSIM$\uparrow$ & LPIPS$\downarrow$ & PSNR$\uparrow$ & SSIM$\uparrow$ & LPIPS$\downarrow$ & PSNR$\uparrow$ & SSIM$\uparrow$ & LPIPS$\downarrow$ & PSNR$\uparrow$ & SSIM$\uparrow$ & LPIPS$\downarrow$\\
				\midrule


                BAD-Gaussians \citep{zhao2024badgaussians} 
                & 24.32 & 0.73 & 0.12 
                & 25.87 & 0.79 & 0.23 
                & 19.06 & 0.62 & 0.19 
                & 23.37 & 0.79 & 0.11 \\

			BAD-Gaussians+bilagrid \citep{wang2024bilateral} 

                & \cellcolor{yellow!25}28.25 & \cellcolor{yellow!25}0.79 & \cellcolor{yellow!25}0.08 
                & \cellcolor{yellow!25}31.99 & \cellcolor{yellow!25}0.86 & \cellcolor{yellow!25}0.06
                & \cellcolor{yellow!25}22.16 & \cellcolor{yellow!25}0.65 & \cellcolor{yellow!25}0.15 
                & \cellcolor{yellow!25}26.48 & \cellcolor{yellow!25}0.81 & \cellcolor{yellow!25}0.09 
                \\

				\specialrule{0.08em}{1pt}{1pt}
                Casual3DHDR-random (ours) 
                & \cellcolor{orange!25}31.20 & \cellcolor{orange!25}0.88 & \cellcolor{red!25}0.05 
                & \cellcolor{orange!25}32.95 & \cellcolor{orange!25}0.87  & \cellcolor{orange!25}0.10 
                & \cellcolor{orange!25}23.65 & \cellcolor{orange!25}0.69 & \cellcolor{orange!25}0.12 
                & \cellcolor{orange!25}29.60 & \cellcolor{orange!25}0.84 & \cellcolor{orange!25}0.05   \\
                
				Casual3DHDR-gt (ours) 
                & \cellcolor{red!25}32.00 & \cellcolor{red!25}0.91 & \cellcolor{orange!25}0.07 
                & \cellcolor{red!25}34.53 & \cellcolor{red!25}0.96 & \cellcolor{red!25}0.05 
                & \cellcolor{red!25}29.35 & \cellcolor{red!25}0.87 & \cellcolor{red!25}0.08 
                & \cellcolor{red!25}33.01 & \cellcolor{red!25}0.93 & \cellcolor{red!25}0.04   \\

				\specialrule{0.08em}{1pt}{1pt}
			\end{tabular}
		}
	\end{center}
\end{table*}

\vspace{0.5em}
We added comparison against the bilateral grid method \citep{wang2024bilateral} applied to \texttt{gsplat} \citep{ye2024gsplatopensourcelibrarygaussian} and BAD-Gaussians \citep{zhao2024badgaussians} in \tabnref{more_table_our_synthesis_deblur} and \tabnref{more_novel_view_synthetic}. The bilateral grids applied to NeRFs and 3DGS gives robustness to large appearance changes, enabling high quality 3D LDR reconstruction and mid-tone rendering quality. However, bilateral grids are not compatible with representing a 3D HDR scene, thus gives degraded renderings on high-contrast, over-exposured and under-exposured views.



\begin{table*}
    \begin{center}
        \captionsetup {font={small,stretch=0.8}}
        \caption{{\textbf{Quantitative comparisons on  \textit{Realsense} and \textit{SmartPhone} dataset under novel view.}}}
        \vspace{-1em}
        \label{more_novel_view_real_table}
        \setlength\tabcolsep{9pt}
        \setlength{\belowcaptionskip}{-10pt}
        \scriptsize
        \resizebox{\linewidth}{!}{
            \begin{tabular}{l|ccc|ccc}
                \toprule
                & \multicolumn{3}{c}{\textbf{Yakitori}} & \multicolumn{3}{|c}{\textbf{Toufu}} \\
                Method & PSNR$\uparrow$ & SSIM$\uparrow$ & LPIPS$\downarrow$ & PSNR$\uparrow$ & SSIM$\uparrow$ & LPIPS$\downarrow$ \\
                \midrule
                gsplat \citep{ye2024gsplatopensourcelibrarygaussian} &25.04 & \cellcolor{yellow!25}0.83 & \cellcolor{yellow!25}0.27 & 29.88 &0.81 & \cellcolor{yellow!25}0.24 \\
                BAD-Gaussians \citep{zhao2024badgaussians} &23.31 &0.78 &0.28 &30.05 &0.82 & \cellcolor{yellow!25}0.24 \\
                HDR-Plenoxels \citep{jun2022hdr}& 27.13  & 0.81 & 0.33 &30.91 &0.82 &0.29 \\
                Gaussian-W \citep{zhang2024gaussian}  &\cellcolor{yellow!25}27.57 & \cellcolor{orange!25}0.84 &0.28 & \cellcolor{yellow!25}30.89 & \cellcolor{yellow!25}0.83 &0.26 \\
                \specialrule{0.08em}{1pt}{1pt}
                Casual3DHDR-random (ours) & \cellcolor{orange!25}28.56 & \cellcolor{orange!25}0.84 & \cellcolor{orange!25}0.22 & \cellcolor{orange!25}32.75 & \cellcolor{orange!25}0.87 & \cellcolor{red!25}0.17 \\
                Casual3DHDR-gt (ours) & \cellcolor{red!25}29.19 & \cellcolor{red!25}0.87 & \cellcolor{red!25}0.16 & \cellcolor{red!25}32.84 & \cellcolor{red!25}0.91 & \cellcolor{orange!25}0.18 \\
                \bottomrule
            \end{tabular}
        }

        \vspace{1.0em}

        \setlength\tabcolsep{1pt}
        \resizebox{\linewidth}{!}{
			\begin{tabular}{l|ccc|ccc|ccc|ccc}
				\toprule
				&  \multicolumn{3}{c}{Fish-pixel8pro}  &  \multicolumn{3}{|c}{Building-pixel8pro}  &  \multicolumn{3}{|c}{Toufu-vicon}  &  \multicolumn{3}{|c}{Girls-vicon} \\
				& PSNR$\uparrow$ & SSIM$\uparrow$ & LPIPS$\downarrow$ & PSNR$\uparrow$ & SSIM$\uparrow$ & LPIPS$\downarrow$ & PSNR$\uparrow$ & SSIM$\uparrow$ & LPIPS$\downarrow$ & PSNR$\uparrow$ & SSIM$\uparrow$ & LPIPS$\downarrow$\\
				\midrule

                gsplat \citep{ye2024gsplatopensourcelibrarygaussian} &23.20 & \cellcolor{yellow!25}0.82 & 0.16 &25.99 &0.81 & \cellcolor{yellow!25}0.11 &24.34 &0.81 &0.28 &23.81 &0.77 &0.28 \\
                gsplat+bilagrid \citep{wang2024bilateral} &25.26 &0.78 &0.14 &25.47  &0.77  &0.16  &30.48 &0.82 &\cellcolor{yellow!25}0.17 &26.76 &0.69 &0.25 \\
				BAD-Gaussians \citep{zhao2024badgaussians} &24.28 &0.78 &\cellcolor{yellow!25}0.14 &26.93 &\cellcolor{yellow!25}0.82 & \cellcolor{yellow!25}0.11 &24.22 &0.82 & 0.24 &23.95 &0.77 &0.28 \\
                BAD-Gaussians+bilagrid \citep{wang2024bilateral} &25.12 & 0.77 & 0.17 &25.63 &0.77 & 0.15 &30.52\cellcolor{yellow!25} &\cellcolor{yellow!25}0.83 &\cellcolor{yellow!25}0.17 &26.18 &0.71 &\cellcolor{yellow!25}0.23 \\
                HDR-Plenoxels \citep{jun2022hdr} &19.39 &0.53 &0.65 &26.87 &0.81 &0.15 &17.90 &0.51 &0.69 &26.73 &0.84 &0.30 \\
                Gaussian-W \citep{zhang2024gaussian}&\cellcolor{yellow!25}26.13 & \cellcolor{orange!25}0.83 &0.15 &\cellcolor{yellow!25}27.99 &\cellcolor{yellow!25}0.82 &\cellcolor{yellow!25}0.11 &\cellcolor{yellow!25}26.38 & \cellcolor{yellow!25}0.83 &0.29 &\cellcolor{yellow!25}26.88 & \cellcolor{yellow!25}0.86 & 0.25 \\
                \specialrule{0.08em}{1pt}{1pt}
                Casual3DHDR-random (ours) &\cellcolor{orange!25}28.30 & \cellcolor{orange!25}0.83 & \cellcolor{orange!25}0.13 &\cellcolor{orange!25}28.79 & \cellcolor{orange!25}0.83 &\cellcolor{orange!25}0.09 &\cellcolor{orange!25}30.87 & \cellcolor{orange!25}0.90 &\cellcolor{orange!25}0.15 &\cellcolor{orange!25}32.00 & \cellcolor{orange!25}0.90 & \cellcolor{orange!25}0.19 \\
				Casual3DHDR-gt (ours) &\cellcolor{red!25}30.81 & \cellcolor{red!25}0.87 & \cellcolor{red!25}0.12 &\cellcolor{red!25}29.71 & \cellcolor{red!25}0.85 &\cellcolor{red!25}0.08 &\cellcolor{red!25}31.34 & \cellcolor{red!25}0.92 &\cellcolor{red!25}0.12 &\cellcolor{red!25}32.39 & \cellcolor{red!25}0.91 & \cellcolor{red!25}0.17 \\
				\specialrule{0.08em}{1pt}{1pt}
			\end{tabular}
		}
    \end{center}
\end{table*}

\begin{figure*}[h]
    \includegraphics[width=\textwidth]{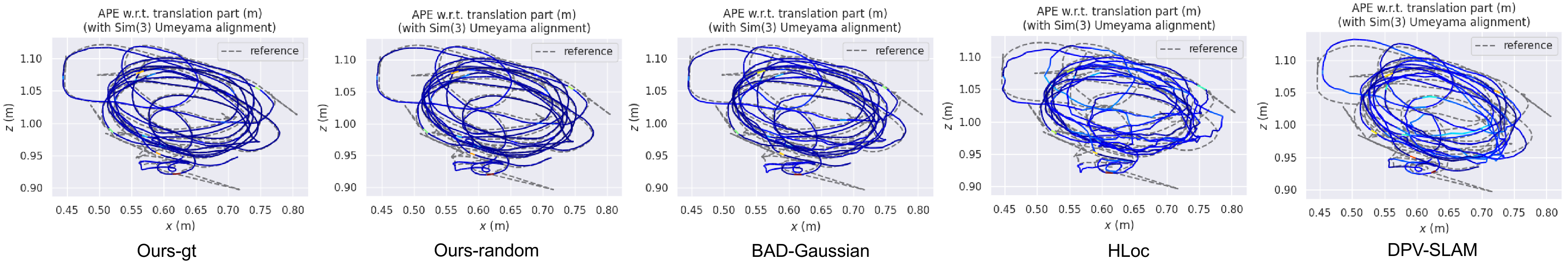}
    \vspace{-2.5em}
    \caption{\textbf{Qualitative comparison for pose estimation on the Girls-vicon sequence of the \textit{Realsense} dataset.}
    }
        \label{girls_traj_quali}
\end{figure*}

\begin{figure*}[h]
    \includegraphics[width=\textwidth]{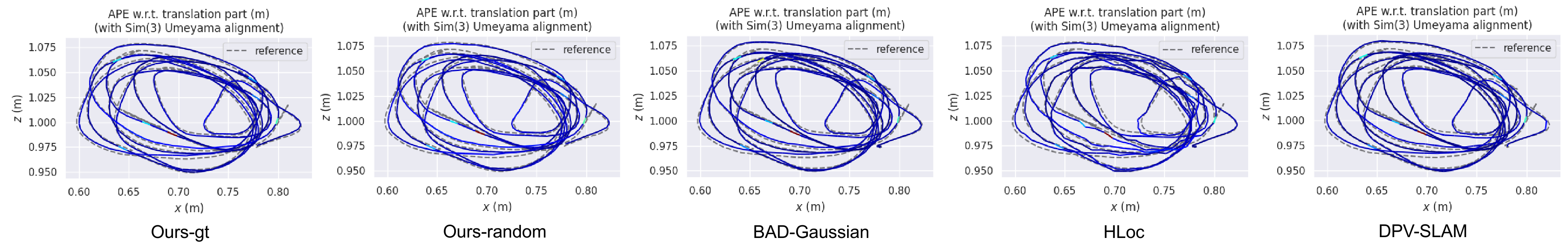}
    \vspace{-2.5em}
    \caption{\textbf{Qualitative comparison for pose estimation on the Toufu-vicon sequence of the \textit{Realsense} dataset.}
    }
    \label{toufu_traj_quali}
\end{figure*}

\subsection{More experimental results under novel view.}
\begin{figure*}
 \centering
 \includegraphics[width=\textwidth]{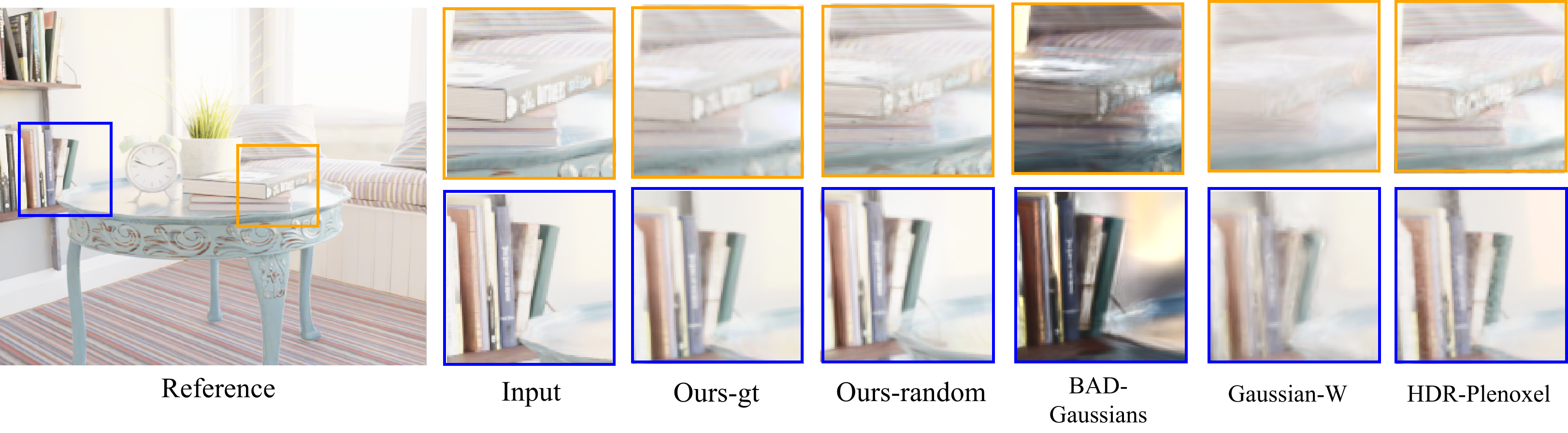}
 \includegraphics[width=\textwidth]
 {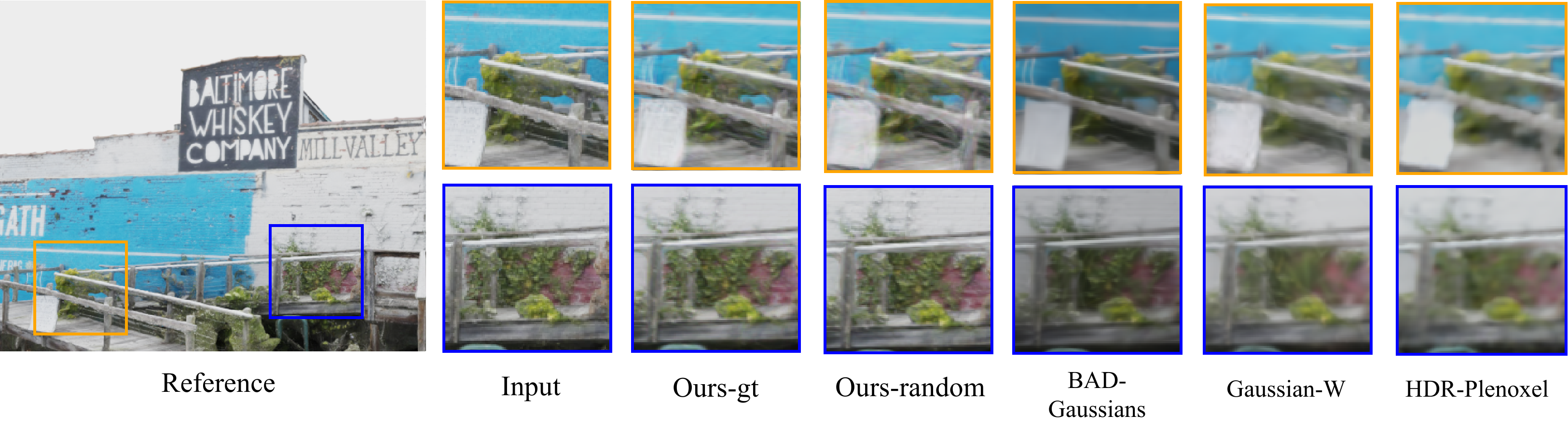}
  \includegraphics[width=\textwidth]
{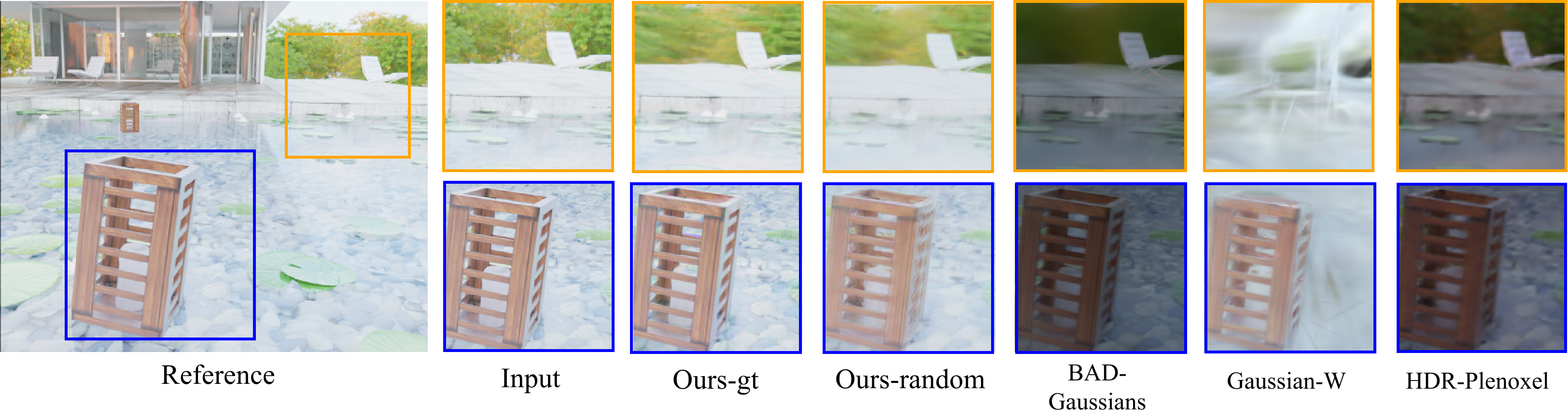}
\vspace{-2em}
\caption{\textbf{Qualitative comparison on \textit{synthetic} dataset(Cozyroom, Factory, Outdoorpool) under novel view.
 }}
	\label{more_novel_view_synthetic}
\end{figure*}

\begin{figure*}
	\setlength\tabcolsep{1pt}
	\centering
	\includegraphics[width=\textwidth]{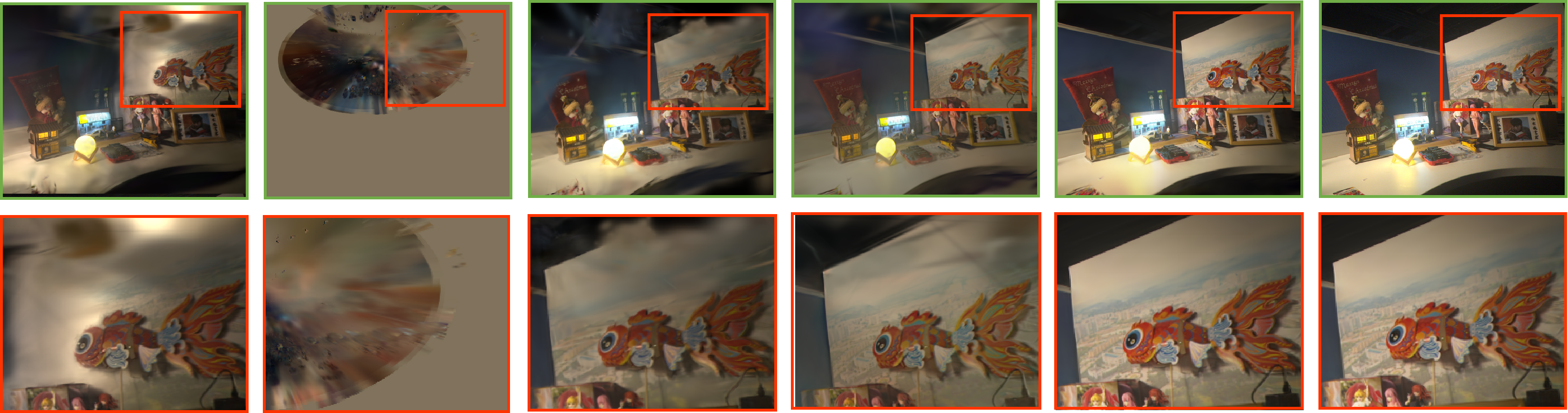}\\ \ \\
 	\includegraphics[width=\textwidth]{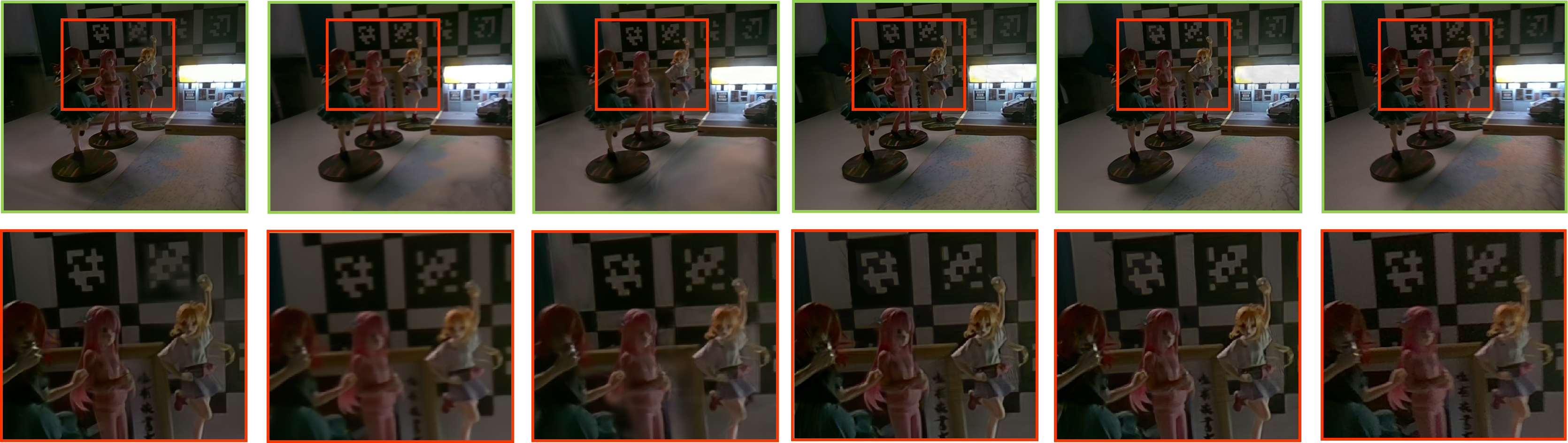}\\ \ \\
  	\includegraphics[width=\textwidth]{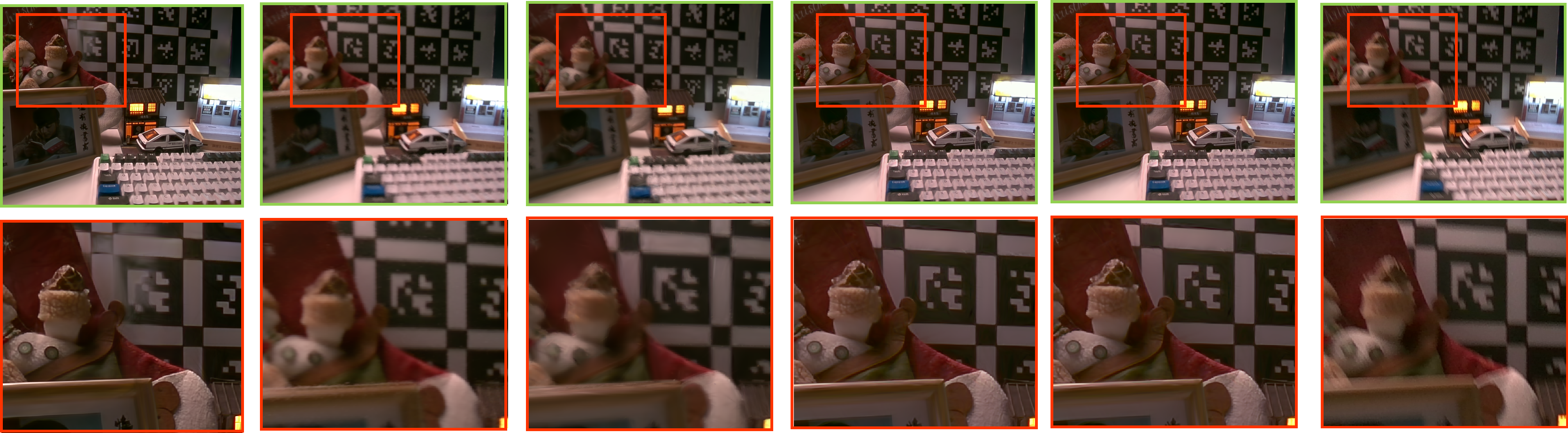}\\ 
	\small{
		\begin{tabular*}{
			\linewidth}{
				@{\extracolsep{\fill}}C{0.162\textwidth}C{0.162\textwidth}C{0.162\textwidth}C{0.162\textwidth}C{0.162\textwidth}C{0.162\textwidth}}
			BAD-Gaussians & HDR-Plenoxels & Gaussian-W & Ours-random & Ours-gt & Reference
		\end{tabular*}
	}
    \vspace{-1.5em}
	\caption{{\textbf{Qualitative comparison on \textit{Smartphone} dataset under novel view.
 }}}
    \vspace{-1.0em}
	\label{more_novel_view_real}
\end{figure*}

The quantitative experimental results in \tabnref{more_novel_view_real_table} indicate that our method significantly outperforms previous approaches in novel view synthesis on two real-world datasets. Further, the qualitative experimental results in \figrefer{more_novel_view_synthetic} and \figrefer{more_novel_view_real} demonstrate that our method produces higher-quality rendered images under novel viewpoints compared to other approaches. These results indicate that our method is capable of learning accurate HDR scene representations and implicit CRF representations.

\begin{figure*}
\includegraphics[width=\textwidth]{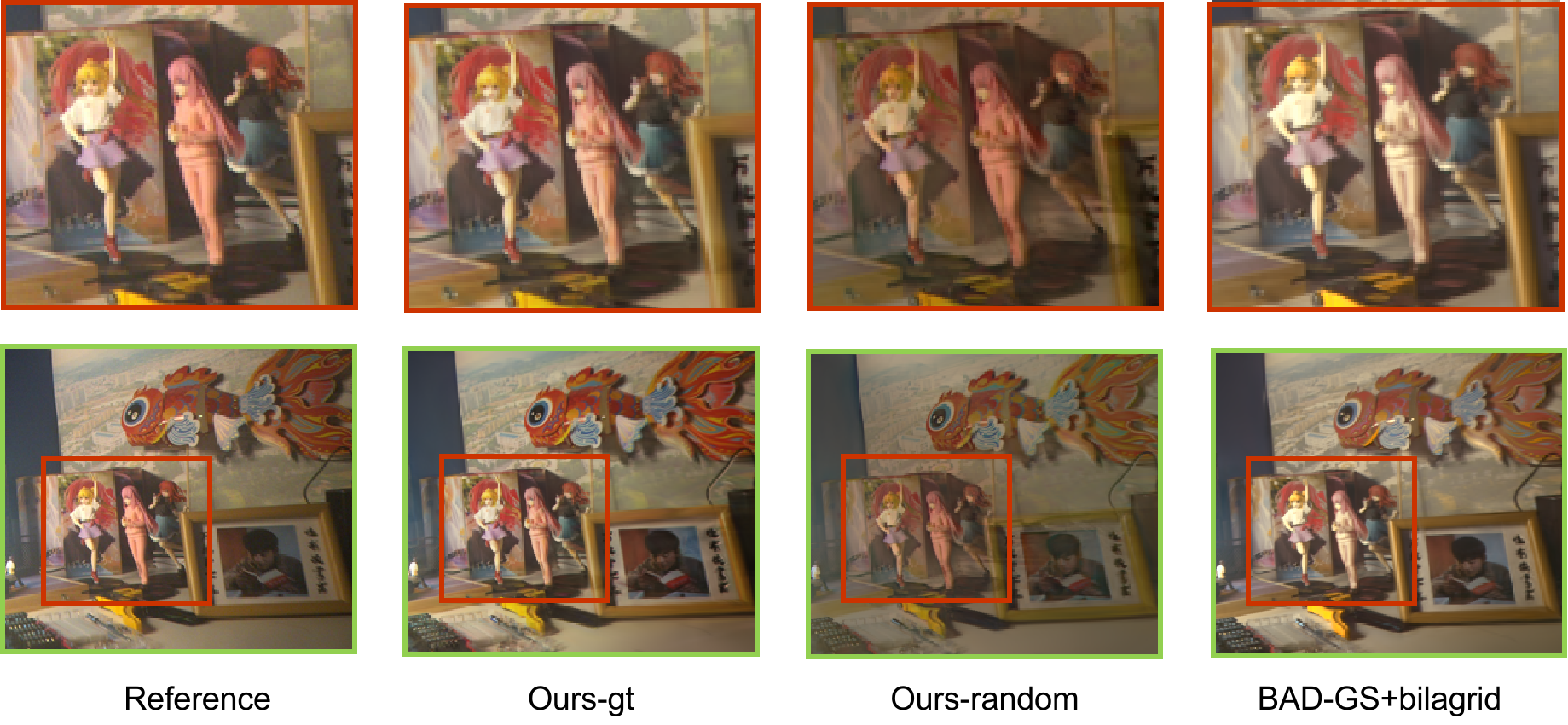}
\includegraphics[width=\textwidth]
{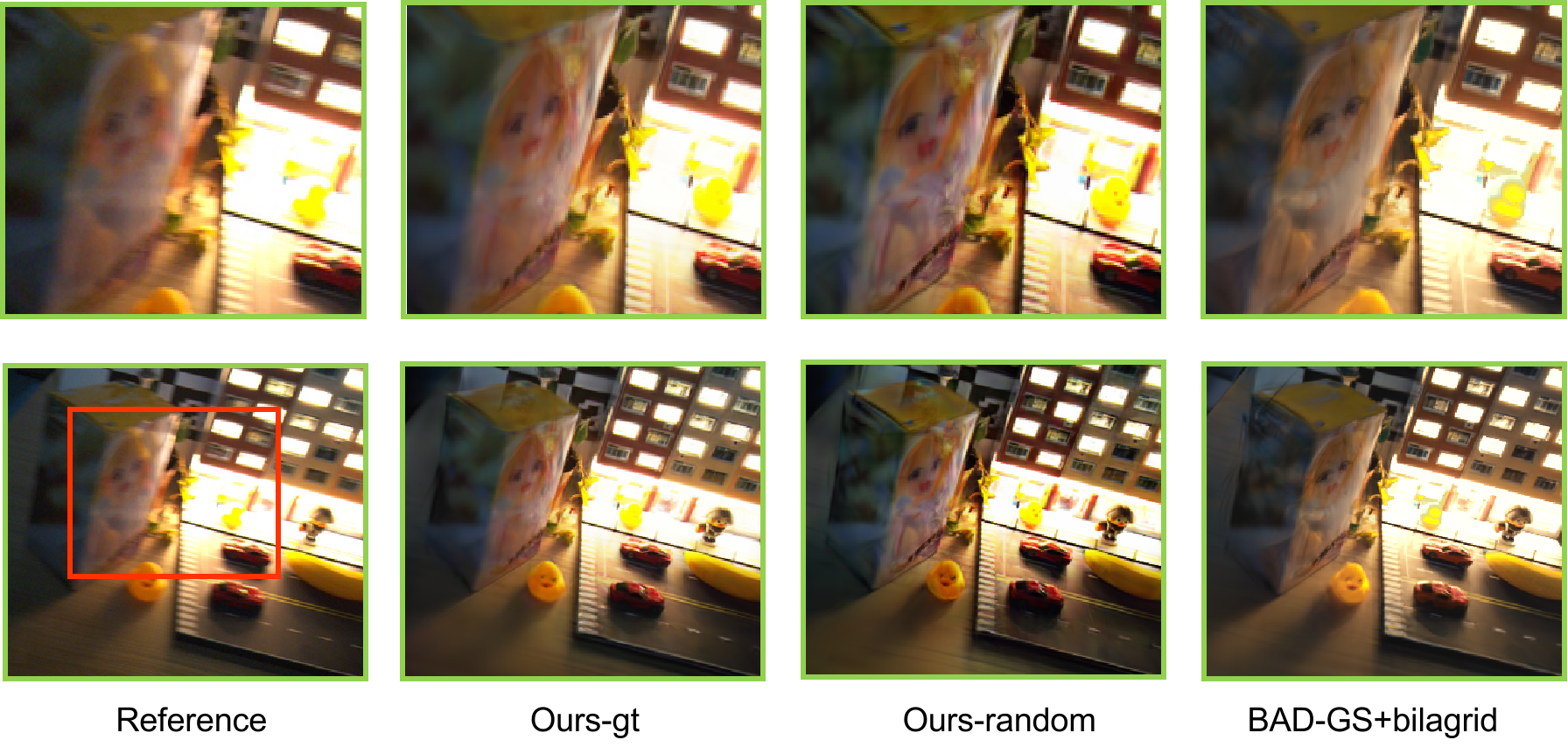}

\caption{\textbf{Qualitative comparison with bilateral method on \textit{Smartphone} dataset under novel view.} 
It is better to view the results on a monitor with high resolution and a gamut coverage close or better than sRGB.}
	\label{bilateral_grid_results}
\end{figure*}

In addition, we compare against the bilateral grid method \citep{wang2024bilateral} applied to \texttt{gsplat} \citep{ye2024gsplatopensourcelibrarygaussian} and BAD-Gaussians \citep{zhao2024badgaussians}, as shown in \tabnref{more_novel_view_real_table} and \figrefer{bilateral_grid_results}. As aforementioned, with bilateral grid \citep{wang2024bilateral}, BAD-Gaussians \citep{zhao2024badgaussians} cannot represent the HDR details of the 3D scenes, thus yields degraded renderings in the high-contrast areas. As it is showed in the \figrefer{bilateral_grid_results}, the girls in the Fish sequence of the \textit{Smartphone} dataset are over-exposed in some views, thus exhibits under-saturation; Meanwhile, the duck in the Building sequence of the \textit{Smartphone} dataset has lost its details and exhibits artifacts on its edge.

\subsection{More experimental results on pose estimation.}

To demonstrate that our method can accurately recover the continuous camera motion trajectory, we visualized and compared the trajectory optimized by our method with the ground truth trajectory, as well as with other baselines. The qualitative results in \figrefer{girls_traj_quali} and \figrefer{toufu_traj_quali} indicate that our method achieves higher pose estimation accuracy compared to previous methods. 


\subsection{Video results of Casual3DHDR}

To show the great performance of our approach, we provide two supplementary videos.

\texttt{novel\_view\_spiral\_traj.mp4} uses the spiral trajectory to render novel views with the given exposure times. The upper half of the video shows the rendered images while the lower half of the video shows the exposure time of current frame, as the camera moves along a spiral trajectory.

\texttt{train\_view\_traj.mp4} uses the trajectory and exposure times of training view. On the lower half of the video, we show that our approach is robust to blurry inputs and can render the sharp images with given exposure times.


\cleardoublepage

\bibliographystyle{ACM-Reference-Format}
\bibliography{acmart}

\end{document}